\documentclass{article}
\usepackage[british]{babel}
\usepackage{jfrExamplee}
\usepackage{graphicx}
\usepackage{setspace}
\usepackage{apacite}
\usepackage{natbib}
\usepackage{epstopdf}
\usepackage{subcaption}

\usepackage{amsmath}
\usepackage{amssymb}
\usepackage{xcolor}
\usepackage{multicol}

\usepackage{algorithm}
\usepackage{algorithmic}

\usepackage{endnotes}
\let\footnote=\endnote

\usepackage{lineno}

\title{Advancement and Field Evaluation of a Dual-arm Apple Harvesting Robot}

\author{
	Keyi Zhu \\
	Department of Mechanical Engineering \\ 
	Michigan State University \\ 
	East Lansing, MI 48824, USA \\
	\texttt{zhukeyi1@msu.edu} \\
	\And
	Kyle Lammers \\
	Department of Electrical and Computer Engineering \\ 
	Michigan State University \\ 
	East Lansing, MI 48824, USA \\
	\texttt{lammer18@msu.edu}
	\And
	Kaixiang Zhang \\
	Department of Mechanical Engineering \\ 
	Michigan State University \\ 
	East Lansing, MI 48824, USA \\
	\texttt{zhangk64@msu.edu} \\
	\And	
        Chaaran Arunachalam \\
	Department of Electrical and Computer Engineering \\ 
	Michigan State University \\ 
	East Lansing, MI 48824, USA \\
	\texttt{arunach6@msu.edu}
	\And
        Siddhartha Bhattacharya \\
	Department of Comp. Sci. and Engineering \\ 
	Michigan State University \\ 
	East Lansing, MI 48824, USA \\
	\texttt{bhatta70@msu.edu}
	\And
        Jiajia Li \\
	Department of Electrical and Computer Engineering \\ 
	Michigan State University \\ 
	East Lansing, MI 48824, USA \\
	\texttt{lijiajia@msu.edu}
	\And
	Renfu Lu \\
	United States Department of Agriculture \\ 
	Agricultural Research Service \\
	East Lansing, MI 48824, USA \\
	\texttt{renfu.lu@usda.gov} \\
	\And
	Zhaojian Li\thanks{Zhaojian Li is the corresponding author.} \\
	Department of Mechanical Engineering \\ 
	Michigan State University \\ 
	East Lansing, MI 48824, USA \\
	\texttt{lizhaoj1@egr.msu.edu} \\
}

\begin{document}
\maketitle


 \vspace{-16pt}
\begin{abstract}
\vspace{-14pt}
Apples are among the most widely consumed fruits worldwide. Currently, apple harvesting fully relies  on  manual labor, which is costly, drudging, and hazardous to workers. Hence, robotic harvesting technology has attracted increasing attention in recent years. However,  existing systems still fall short in terms of harvesting performance, cost effectiveness, and reliability for operating in complex orchard environments. 
In this work, we present the development and evaluation of a dual-arm apple harvesting robot. The system integrates a Time-of-Flight (ToF) camera, two 4-degree-of-freedom robotic arms, a centralized vacuum system, and a fruit post-harvest handling module. During harvesting, suction force is dynamically assigned to either arm via the centralized vacuum system, enabling efficient apple detachment while significantly reducing power consumption and noise.
Compared to our previous design, we incorporated a platform movement mechanism that enables both in-out and up-down adjustments, enhancing the robot’s dexterity and adaptability to varying canopy structures. On the algorithmic side, we developed a robust apple localization pipeline that combines a foundation-model-based detector, pixel-wise segmentation, and clustering-based depth estimation, which improves performance in outdoor conditions. Additionally, pressure sensors were integrated into the vacuum system, and a novel dual-arm coordination strategy was introduced to respond to harvest failures based on sensor feedback, further improving picking efficiency.
Field demonstrations were conducted in two commercial orchards in Michigan, USA, with different canopy structures. The system achieved success rates of 80.7\% and 79.7\%, with an average picking cycle time of 5.97 seconds. The proposed coordination strategy reduced harvest time by 28\% compared to a single-arm baseline.
The proposed dual-arm harvesting robot enhances the reliability and efficiency of apple picking. With further advancements in hardware and software, the system holds strong potential for fully autonomous operation and future commercialization to support the apple industry.

\end{abstract}

\section{Introduction}\label{sec:intro}
Apple production is a major sector within agriculture but faces mounting challenges due to rising labor costs, shortage of skilled workers, and safety risks associated with manual harvesting. It is estimated that fruit harvesting accounts for approximately 10.4\% of the total apple production cost~\citep{gallardo20202019}. Fruit picking involves performing repetitive and physically demanding tasks, raising concerns regarding human safety. These issues underscore the urgent need for automated harvesting solutions to reduce both economic burdens and health risks.
As orchards have been increasingly transitioned from unstructured layouts to more structured systems, such as V-trellis and vertical fruiting walls, they also create a more suitable environment for robotic harvesting, accelerating the shift toward automation and helping to further offset labor challenges.

Fruit harvesting robots have received significant research attention over the past decades~\citep{rajendran2024towards,zhou2022intelligent}. Robotic harvesting systems have been developed for a variety of crops, including apples~\citep{silwal2017design,brown2021design,wang2022geometry,li2023multi,zhang2024automated}, kiwifruit~\citep{williams2019robotic,barnett2020work}, sweet peppers~\citep{lehnert2017autonomous,arad2020development}, strawberries~\citep{xiong2020autonomous}, and tomatoes~\citep{gao2022development}.
These systems are generally categorized into two types. The first type is shake-and-catch robots~\citep{zhang2020field}, which detach fruits by shaking the main trunk or branches, allowing the fruits to fall into a collection module. While this method enables high harvesting efficiency, it can cause mechanical damage to trees and bruising to fruits due to impact.
The second type is fruit-by-fruit harvesting~\citep{ren2024mobile,li2023multi}, where robotic manipulators are used to pick individual fruits in a controlled manner. This approach minimizes damage to both fruit and trees, making it more suitable for high-quality harvesting. However, its efficiency remains a key challenge, particularly when compared to manual labor under time-sensitive harvest conditions. 
Generally, a fruit-by-fruit harvesting robot consists of three key modules:
(1) a perception module, which detects and localizes target fruits;
(2) a picking module, which controls the robotic arm to harvest the localized fruit; and
(3) a collection module, which gathers the picked fruits for storage and transportation.
Numerous fruit harvesting robots have been developed with various mechanical designs and algorithmic approaches. For instance,~\cite{lehnert2017autonomous} designed a system using a manipulator, from Universal Robots (UR), equipped with an eye-in-hand RGB-D camera to scan sweet peppers and determine grasping poses. The robot uses a suction cup to attach to the fruit’s surface, followed by stem cutting using a blade. In~\cite{ren2024mobile}, a UR3e-based system employs a soft gripper that performs a drag-then-rotate motion to detach strawberries without causing damage. Similarly,~\cite{yin2023development} proposed a clamp-shaped end-effector for citrus harvesting, where integrated blades sever the stem as the clamp secures the fruit. However, picking efficiency remains an issue for these robots, as the average cycle time for each harvest attempt exceeds 10 seconds.

Specific to apple harvesting, several robots have also been developed. In~\cite{silwal2017design}, the authors built a 6-degree-of-freedom manipulator equipped with a tendon-driven end-effector for fruit detachment, but their image processing algorithm relies on controlled background, limiting the adaptability in practical scenarios. \cite{bu2022design} introduced a claw-like gripper combined with various path planning strategies for efficient apple picking, however, the robot suffers from the efficiency problem, with the cycle time greater than 10 seconds for each grasping pattern. In~\cite{wang2022geometry}, a UR5 robot with a soft end-effector was used, alongside a novel end-to-end neural network that directly infers grasping poses from RGB-D images. Additionally,~\cite{li2023multi} developed a multi-arm robotic system and applied multi-agent reinforcement learning for inter-arm coordination, achieving higher harvesting efficiency than single-arm configurations. Despite the promising performance they have achieved, the multiple cameras installed on the robots post high requirement on the computational units, which limits their applicability in commercial scenarios. Also, the evaluations of the robots in commercial orchards remain limited, and current systems still fall short of the speed and accuracy required for practical deployment.

The main challenges associated with fruit-by-fruit harvesting robots are two-fold.
First, a fast and robust perception algorithm is essential, as the accuracy of fruit localization directly affects the robot’s harvesting success rate. Compared to indoor manipulation tasks, fruit perception in outdoor orchard environments is significantly more complex. The canopy structure is often irregular, with fruits partially or fully occluded by foliage, branches, or support structures, making reliable detection and localization more difficult ~\cite{li2023multi}. Moreover, variable lighting conditions—such as backlighting and overexposure—can distort fruit color and introduce significant noise into depth images, particularly under intense sunlight. In addition, fruit clustering poses further challenges, often leading to incorrect detections or merged instances. Such errors can result in harvesting failures or physical damage to the fruit ~\cite{bu2022design}.
Second, picking efficiency is of critical concern. The harvest window for fruits is typically short, placing high demands on the robot’s operational speed. If the robot fails to harvest fruits in a timely manner, fruit quality may decline, ultimately reducing market value and economic viability. In addition, the cost-effectiveness of a robotic harvesting system highly depends on its harvesting speed. Therefore, improving cycle time is key to commercial viability of robotic harvesters ~\cite{lehnert2017autonomous,bu2022design,ren2024mobile}.

To address the aforementioned challenges, our group has been developing an automated apple harvesting robot, aiming for various orchard environments with a picking rate competitive to that of human labor. Beginning with a single-arm configuration~\citep{zhang2021system,zhang2022algorithm,zhang2024automated}, we transitioned to a dual-arm setup in 2023~\citep{lammers2024development}. The current robot features two tube-shaped robotic arms, each with 4 degrees of freedom, powered by a centralized vacuum system. Equipped with specially designed soft end-effectors, the arms can securely attract apples with minimal air leakage, enhancing the harvest success rate. A single custom-designed vacuum system with an automated valve control dynamically distributes suction force between the two arms. Compared to conventional gripper-based systems, our vacuum-driven approach reduces cycle time, improves tolerance to localization errors, and minimizes fruit bruising.

Built on our previous dual-arm design~\citep{lammers2024development}, several enhancements have been made. On the hardware side, we upgraded the valve control strategy to prevent overheating during extended operation and introduced a platform movement mechanism that repositions the robot to maximize apple accessibility with a greater workspace.
On the software side, we developed a new perception pipeline incorporating a foundation-model-based apple detector, pixel-wise segmentation, and clustering-based depth estimation. This approach significantly improves perception robustness under challenging outdoor conditions, such as canopy occlusion and extreme lighting. Compared to our previous work~\citep{chu2021deep,chu2023o2rnet}, the new algorithm delivers more reliable detection and localization. Furthermore, it offers performance comparable to our prior active laser-scanning method~\citep{chu2024high}, while substantially reducing localization time.
Additionally, we introduced a dual-arm coordination strategy that utilizes real-time pressure feedback from the valve system to infer picking status and react to harvest failures. This reactive strategy enables greater independence between the arms, thereby improving overall picking efficiency.

The main contributions of this paper are summarized as follows:
\begin{itemize}
    \item \textbf{Robust Apple Localization Algorithm:} We propose a novel localization pipeline that combines a foundation-model-based apple detector, a custom segmentation model, and a clustering-based depth estimation algorithm. This approach enables reliable apple localization in orchard environments where apples may be partially occluded by canopy structures or affected by extreme lighting conditions such as overexposure.
    \item \textbf{Dual-Arm Coordination Strategy}: We introduce a new coordination strategy based on the Temporal Logic framework, which allows the generation of formal coordination policies for complex tasks. This strategy enables parallel operation of the dual arms even when sharing a single vacuum source. It is also reactive to harvest failures, as detected by real-time pressure feedback from the vacuum system, thereby maintaining high picking efficiency.
    \item \textbf{Comprehensive Field Evaluation}: We conducted systematic field tests in two commercial orchards in Michigan, USA, during the 2024 harvest season. The evaluation demonstrated the effectiveness and superiority of our proposed methods under commercial harvesting conditions.
\end{itemize}

The remainder of this paper is organized as follows. Section~\ref{sec:hardware} describes the hardware system of the robot, including the components used, the vacuum-based harvesting mechanism, and the fruit gathering system we developed and fabricated. Section~\ref{sec:software} presents the software architecture, including the perception algorithm, single-arm control strategy, dual-arm coordination method, and a graphical user interface designed to support in-field testing. Section~\ref{sec:experiment} provides field evaluation results from the 2024 harvest season, along with a performance analysis and failure case review. Section~\ref{sec:discussion} discusses the current limitations of the system and outlines potential directions for improvement. Finally, Section~\ref{sec:conclusion} summarizes the key findings and further work needed.
\begin{figure}[!b]
    \centering
    \includegraphics[width=0.75\linewidth]{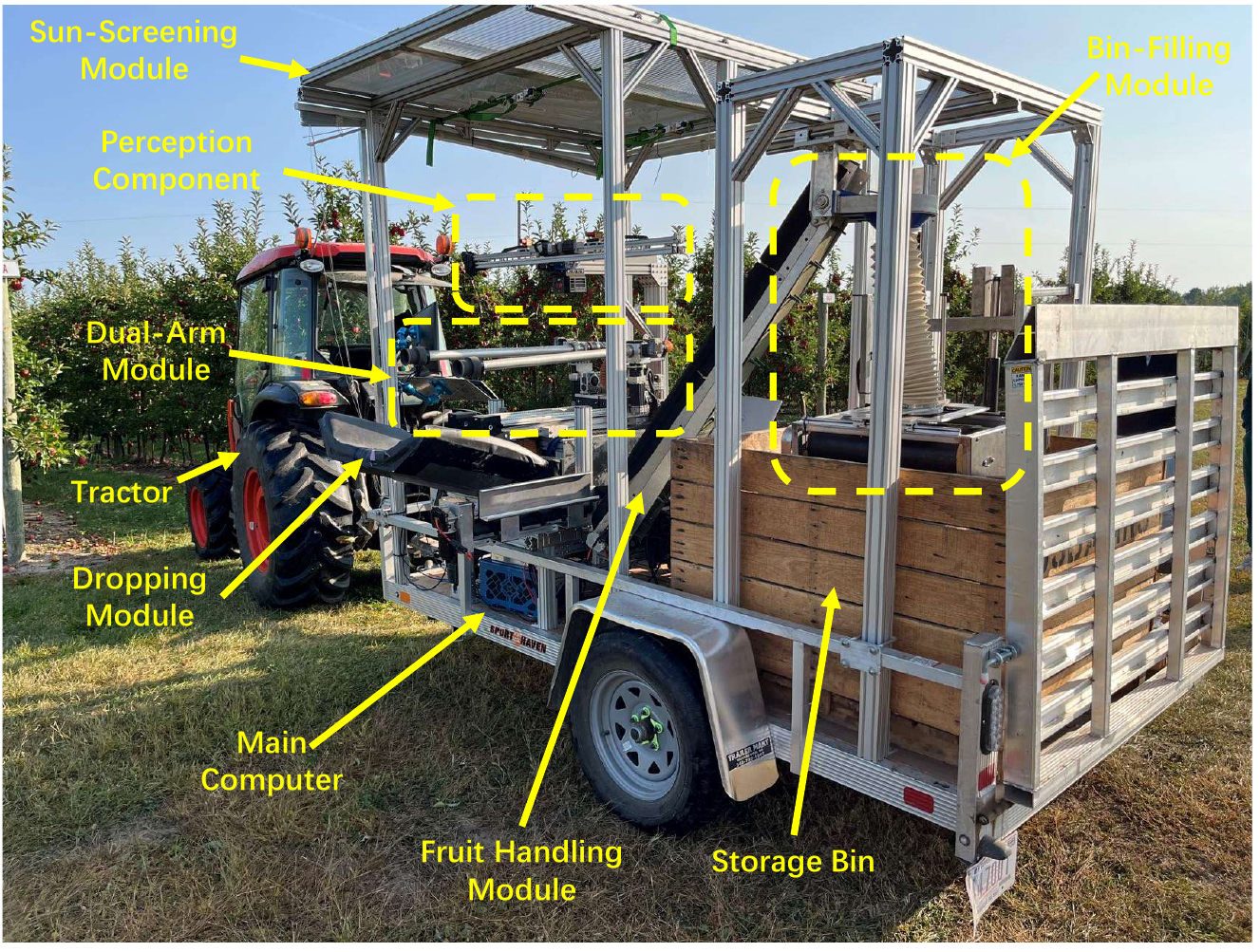}
    \caption{Overview of the dual-arm apple harvesting robot. The system is mounted on a trailer designed to be compatible with most commercially available tractors, allowing for easy transportation and deployment within orchard environments.}
    \label{fig:robot_overview}
\end{figure}

\section{Hardware System}\label{sec:hardware}
The mechanical components of the dual-arm robotic harvesting system are shown in Fig.~\ref{fig:robot_overview}. The robot is mounted on a trailer and can be transported through the orchard using a tractor suitable for uneven terrain. During operation, the robot follows a stop-and-go harvesting pattern controlled by a human operator.

The main functional units of the robot include: (1) a perception component, which captures orchard images for fruit detection and localization; (2) a dual-arm harvester, with each arm using a tube-shaped structure to detach apples via suction from a centralized vacuum system; and (3) an apple gathering system, which receives the detached apples and transfers them to a storage bin without causing bruising. 
In addition to the core components, the robotic system includes several auxiliary modules.
The supporting module includes a Honda gas-powered generator, which provides 240V at 5.5 kW and can power the entire system for over 5 hours on a full tank. This ensures uninterrupted operation during typical harvesting sessions. The operation module consists of a high-performance industrial computer, a monitor, and a mouse and keyboard interface. The computer is equipped with a 24-core Intel® Core™ i9-13900E CPU, 64 GiB of RAM, and two GPUs: an NVIDIA RTX A6000 (48 GiB) and an NVIDIA GeForce RTX 4070 (16 GiB). This setup provides sufficient computational capacity to support all system operations, including motor and sensor communication, real-time execution of perception, planning, and control algorithms, ensuring smooth and responsive performance in the field. A sun-screening module is installed on top of the trailor to reduce direct sunlight interference, improving the reliability of the perception algorithm under harsh lighting conditions.


\subsection{Perception Component}\label{sec:tof}
Accurate perception—including both detection and localization—is critical to successful robotic harvesting. A reliable perception system forms the foundation for robust and efficient performance. To meet this requirement, a Time-of-Flight (ToF) camera, manufactured by Percipio Inc. is employed. The selected model, TL460-S1-E1, has a working range of 0.3--9.5~m and offers a field of view (FOV) of $62^\circ$ horizontally and $49^\circ$ vertically. The depth accuracy is specified to be $\pm4$~mm plus 0.25\% of the measured depth. The camera provides a depth resolution of $640 \times 480$ and an RGB resolution of $1920 \times 1080$. It measures $140~\text{mm} \times 94~\text{mm} \times 70~\text{mm}$ and weighs approximately 1.1~kg. This ToF camera delivers high-quality spatial information of the orchard environment, supporting the robust performance of the perception algorithms.

Unlike eye-in-hand configurations where the camera is mounted on the end-effector and moves with the robotic arm, the ToF camera in our system is fixed on the robot frame, positioned approximately 1~m away from the front edge of the arms’ workspace. This static placement ensures that the camera consistently covers the entire operational workspace of both arms throughout the harvesting process. Moreover, mounting the camera in a fixed position enhances the stability of the captured RGB and depth data by avoiding motion-induced blur and reducing depth measurement errors associated with camera movement.

The primary challenges for perception in outdoor orchard environments are: (1) extreme lighting conditions, and (2) occlusions caused by canopy elements such as foliage and branches. The first issue is addressed by a sun-screening module installed on top of the trailer, as shown in Fig.~\ref{fig:robot_overview}. This module can be extended before harvesting begins, effectively blocking direct sunlight from reaching the canopy and improving the reliability of the perception system. The second challenge is mitigated by our newly developed localization algorithm, which will be detailed in the following section.

\subsection{Dual-Arm Robotic Harvester}\label{sec:robot}
\begin{figure}[htbp]
    \centering
        \centering
        \includegraphics[width=0.5\linewidth]{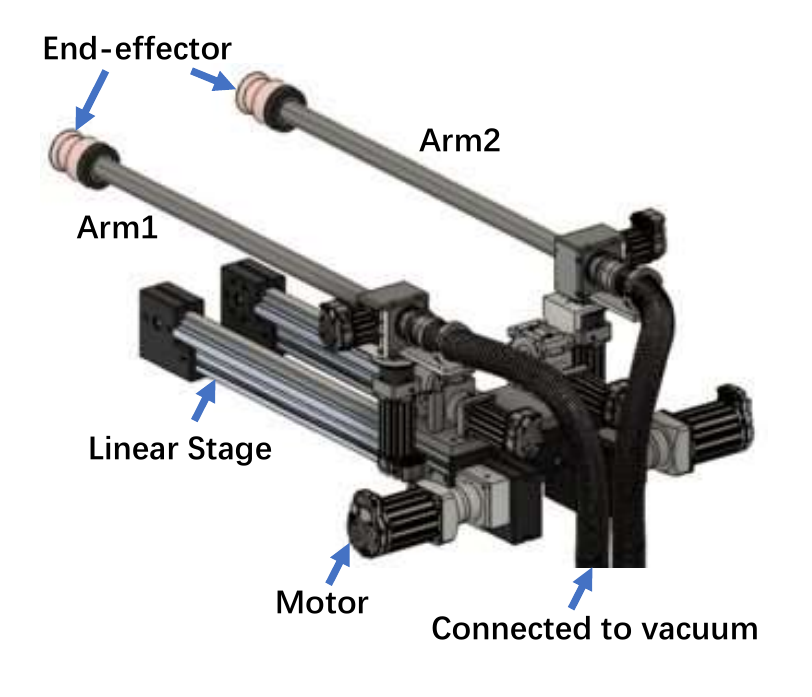}
\caption{The CAD model of the dual-arm manipulator design.}
        \label{fig:dual_arm_cad}
\end{figure}

In our previous work~\citep{zhang2022algorithm}, we developed a 4-degree-of-freedom (DOF) manipulator consisting of one prismatic joint and three revolute joints. Built on that design, our current dual-arm harvesting robot integrates two such 4-DOF manipulators, as illustrated in Fig.~\ref{fig:dual_arm_cad}. Each manipulator features a hollow tube structure that serves as an airflow channel. When connected to the centralized vacuum system, this configuration enables the arm to detach apples using suction force. The two manipulators are positioned side by side with only a 7~mm gap between them, maximizing the overlap in their workspaces. This overlapping workspace is important for enhancing picking efficiency. It minimizes idle time by reducing instances where one arm is inactive due to a lack of reachable apples while the other is still harvesting. Such coordination challenges are common in dual-arm configurations with non-overlapping workspaces.

All joints are actuated by DC servo motors manufactured by Teknic Inc. (Victor, NY, USA), each equipped with an integrated controller and encoder. These motors can achieve a maximum end-effector speed of 0.7748 m/s while carrying a 1.5 kg payload. Communication between the main computer and the motors is established through a Teknic USB communication hub. By implementing parallel I/O operations, the system maintains a communication frequency above 200~Hz, enabling accurate and responsive low-level control of the robotic arms.


A soft end-effector is mounted at the front end of each arm to perform apple detachment. The end-effector is fabricated from silicone rubber with a hardness of 40 Shore A, providing a compliant and secure interface with the fruit. A custom 3D-printed adapter ensures a tight, non-slip connection between the end-effector and the arm. Indoor tests show that, when paired with the Delfin vacuum machine, the end-effector can generate an average suction force of 47~N—sufficient to reliably detach and hold an apple in most harvesting scenarios.

The advantages of our vacuum-based robotic design are two-fold. First, compared to end-effectors that rely on fingers or blades~\citep{li2023multi,gursoy2023towards}, our design imposes fewer demands on the perception algorithm. It only requires a simple 3D position of the target apple for picking and can tolerate localization errors of up to $\pm$1.5~cm, as verified through indoor tests. In contrast, finger- and blade-based designs typically require precise estimation of both the fruit’s position and orientation to align with the stem and perform accurate cutting. This increases the computational burden on the perception module and places stricter constraints on motion planning, which can reduce harvesting efficiency.
Second, unlike other vacuum-based designs in which apples are sucked entirely into the vacuum tube and stored in an internal chamber, our design employs a smaller end-effector that minimizes energy consumption. This approach also significantly reduces the risk of unintended suction of leaves or branches, thereby mitigating common failure modes such as vacuum blockage and fruit damage.

To accommodate varying orchard structures, a platform movement module was designed to adjust the position of the robotic system. This module is actuated by two stepper motors and allows motion in both horizontal and vertical directions. It communicates with the main computer through an Arduino Uno board. The module provides real-time positional feedback to the main computer, enabling the operator to flexibly reposition the platform and maximize the number of apples within the robot's workspace. Moreover, the feedback-enabled design lays the groundwork for future development of fully autonomous platform positioning.

\subsection{Vacuum \& Valve System}\label{sec:valve_system}
To optimize space usage and reduce energy consumption, the apple harvesting robot uses a single vacuum source shared between both arms. The system employs a Delfin industrial vacuum machine (model 202 DS), which features two high-performance motors and delivers a peak power of 5.5~HP.
During harvesting, the vacuum machine operates continuously, while a custom-designed valve system dynamically distributes airflow to the appropriate arm. This configuration provides sufficient suction force—up to 2500mmH\textsubscript{2}O at an airflow rate of 360m\textsuperscript{3}/h—to detach apples effectively throughout the dual-arm operation.

\begin{figure}[htbp]
    \centering
    \includegraphics[width=0.65\linewidth]{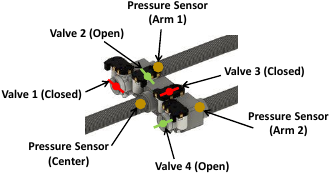}
    \caption{Configuration of the valve system. Valves 1 and 2 control airflow to Arm 1, while Valves 3 and 4 serve Arm 2. In the configuration shown, vacuum flow is directed to Arm 1, enabling it to attach apples, while Arm 2 is open to the atmosphere, allowing any attached apple to be released. Three pressure sensors are integrated into the valve system to monitor the vacuum status in real time.}
    \label{fig:valve}
\end{figure}

To fully leverage the dual-arm configuration and accelerate the harvesting process, we designed a custom valve system, illustrated in Fig.~\ref{fig:valve}. This system dynamically distributes vacuum flow from the centralized vacuum source to either or both robotic arms, enabling coordinated fruit detachment.
The valve system interconnects the vacuum source, the two arms, and the atmosphere, and operates using four butterfly-style gates to control airflow routing. By adjusting the gate states, each arm can independently transition between harvesting and releasing modes. The gates are actuated by compact 12-volt DC motors, with an actuation time of less than one second, ensuring rapid and responsive control of the vacuum distribution and minimizing delays during harvesting.

Pressure sensors are integrated into the valve system to monitor suction conditions and provide feedback to the main computer, further supporting efficient apple picking. Communication between the valve system and the central control unit is handled via Arduino Uno micro-controllers. To address potential thermal issues, a pulse-width modulation (PWM) strategy is employed to regulate motor power, allowing the valve system to operate continuously for over 10 hours without overheating. The Delfin vacuum source is directly connected to the valve system, enabling flexible and efficient vacuum allocation to one or both arms as needed.

The detailed design of the valve system is illustrated in Fig.~\ref{fig:valve}, which incorporates four valves. Valves 1 and 2 control the attachment and release functions of Arm 1, while Valves 3 and 4 serve the same functions for Arm 2. To monitor system status, pressure sensors are installed at three key locations: inside Arm 1, inside Arm 2, and at the vacuum source. By analyzing pressure readings from these sensors, the system can determine whether an apple is successfully attached to the end-effector.

\subsection{Fruit Handling System}\label{sec:fruit_handling}
\begin{figure}[!htbp]
    \centering
    \includegraphics[width=0.6\linewidth]{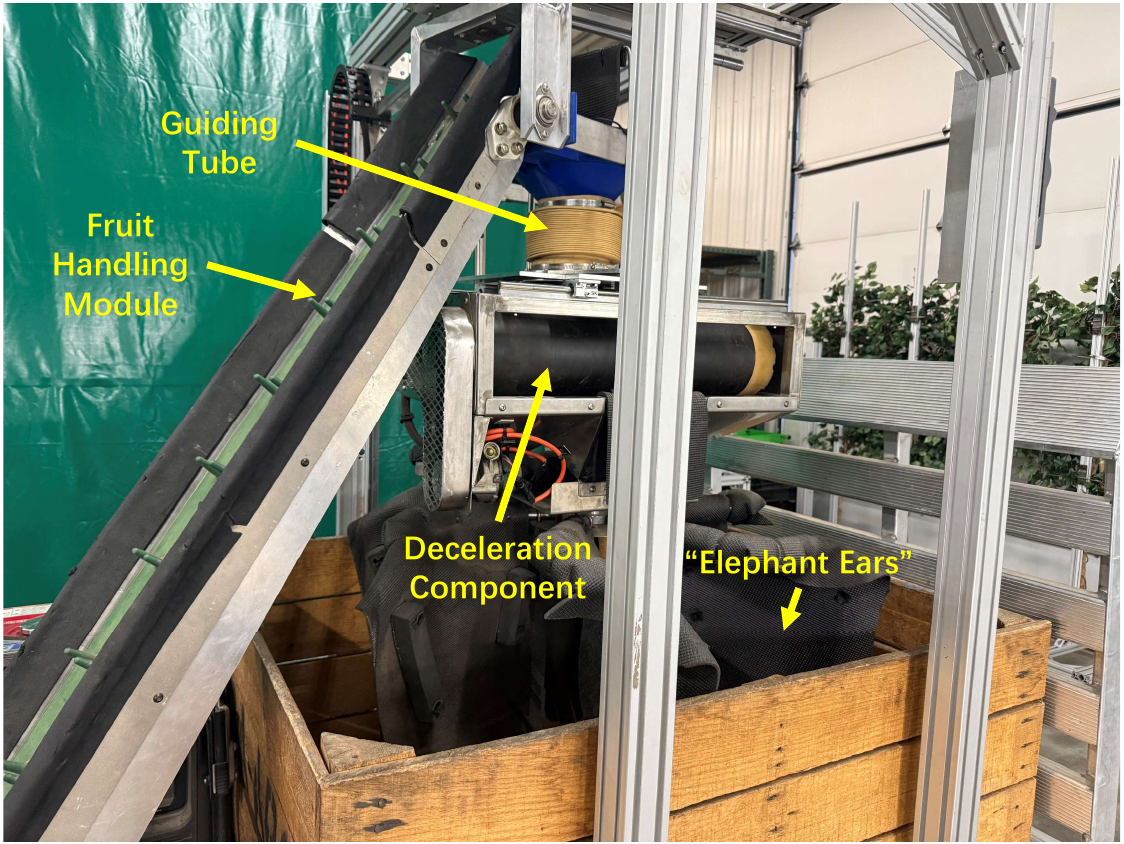}
    \caption{Partial view of the fruit gathering system. The dropping module and storage bin are not shown here due to spatial constraints but are depicted in Fig.~\ref{fig:robot_overview}.}
    \label{fig:fruit_gathering}
\end{figure}

The apple harvester is equipped with a fruit Handling system that facilitates the smooth and gentle transfer of apples for from the robotic harvester to the fruit bin, which is used for either short- or long-term post-harvest storage. As partially shown in Fig.~\ref{fig:fruit_gathering}, the system comprises four main components: a dropping module, a fruit handling module, a bin-filling module, and a storage bin.

The dropping module, shown in Fig.~\ref{fig:robot_overview}, is a sloped surface covered with soft foam. It is positioned to receive apples released by the two robotic arms near their home positions. The foam surface serves to cushion the impact during the drop, minimizing potential bruising or damage to the fruit. Rather than placing the apples gently, this controlled drop method improves harvesting efficiency without causing bruise damage to harvested fruits.
After the apple lands on the dropping module, the slope guides the fruit to the transporting module, which is made up of two specially designed screw and finger conveyors.~\cite{lu2022development} This module transports the apples to the bin-filling system, where they are gently deposited into a storage bin. 

The bin-filling module consists of a soft guiding tube, a deceleration component, and a set of foam panels referred to as ``elephant ears''. When an apple is delivered from the fruit conveying module, it descends freely inside an expandable soft rubber tube to the deceleration component, which slows and controls the apple's movement using a pair of soft cylindrical foam rollers.~\cite{zhang2017development} The decelerated fruit is then guided to the ``elephant ears'', which serve as cushioning pads to avoid the apples coming out from the decelerating foam rollers from contacting or colliding with those apples that are already in the bin, thus preventing fruit bruising.
As the bin fills, the vertical position of the bin-filling system can be adjusted manually or automatically to maintain a proper distance from the surface of the collected apples. This reduces the fruit damage due to collisions. Once the bin is full, the filling module raises to its maximum height, allowing the filled bin to be  removed and replaced with an empty one.

During harvesting, the communication between this system and the main computer is managed through an Arduino micro-controller, allowing the computer to control the system using a simple on/off mode.

\section{Software Design}\label{sec:software}

\subsection{Foundation-Model-Based Apple Localization}\label{sec:perception}
Accurate and efficient perception is a critical component in the loop of autonomous apple harvesting. In complex orchard environments, apples are frequently occluded by foliage, branches, support structures, and other apples. Additionally, extreme lighting conditions—such as direct sunlight or canopy-induced shadows—can cause over-exposure and degrade image quality. These challenges necessitate a robust perception pipeline capable of reliable detection and localization under diverse real-world scenarios.

In our previous work~\citep{chu2021deep}, we developed a novel neural network structure to cope with the occlusion problem. Specifically, we proposed a suppression end after a segmentation network, such as Mask-Region-based Convolutional Neural Networks (Mask-RCNN), to achieve superior performance on our dataset, which was collected in the orchard environment with many occlusion scenarios. We also developed an active laser-scanning-based localization strategy~\citep{chu2024high}, 
obtain a more accurate localization information for target fruits. However, the laser scanning procedure needs extra time, undermining the picking efficiency. Thus, a faster yet reliable localization solution is required. 
Our proposed perception system integrates state-of-the-art deep learning techniques with 3D geometric reasoning. The pipeline consists of three main stages: object detection, segmentation-based pixel identification, and 3D localization through point cloud clustering.

\subsubsection{Object Detection with Foundation Models}\label{sec:detection}
Conventional rule-based object detection methods struggle with generalization and robustness in unstructured environments. In contrast, modern deep learning-based detectors~\citep{ren2016faster,wang2024yolov9} have demonstrated superior performance by learning representative features directly from large-scale datasets. In our system, we employ Grounding-DINO~\citep{liu2023grounding}, a state-of-the-art Transformer-based detector that combines DINO~\citep{zhang2022dino} with Grounded Language-Image Pre-training (GLIP)~\citep{li2022grounded}. Grounding-DINO leverages both visual data and class names during training, allowing it to achieve high accuracy even in occluded or over-exposed scenes.

The network structure of Grounding-DINO is shown in the left part of Fig.~\ref{fig:network_structure}.
To enhance the detection performance and also accelerate the training process, we applied transfer learning by fine-tuning a pre-trained Grounding-DINO model—originally trained on large-scale detection datasets such as COCO, O365, and OpenImage—using our custom orchard dataset. A detection confidence threshold of 0.3 was used to balance recall and precision. Qualitative results demonstrate that the model can reliably detect apples in cluttered scenes without significant false positives.

\subsubsection{Instance Segmentation for Pixel-Level Apple Identification}\label{sec:segmentation}
\begin{figure}[htbp]
    \centering
    \includegraphics[width=0.75\linewidth]{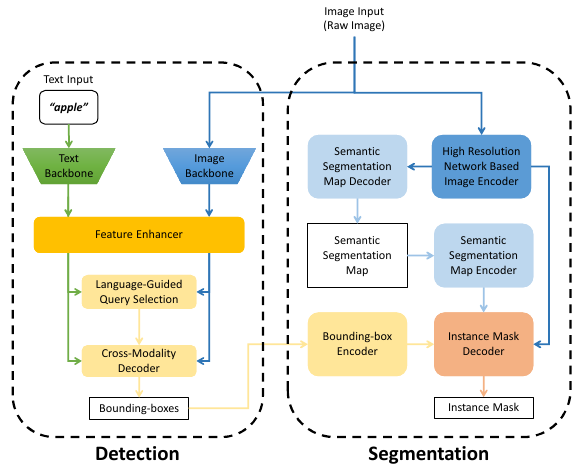}
    \caption{Network architecture of the detection and segmentation pipeline. The detection model (Grounding-DINO) receives both text and image inputs. After feature extraction and cross-modal feature enhancement, it generates bounding boxes corresponding to regions specified by the text input. These bounding boxes are then passed to the segmentation model. The segmentation module performs semantic segmentation based on encoded image features and, using the bounding box information, produces instance masks. These masks effectively separate overlapping objects of the same class in the final output.}
    \label{fig:network_structure}
\end{figure}

With the detected bounding boxes given by Grounding-DINO, we developed a segmentation module capable of both semantic and instance segmentation.
Initially, apple localization relied on the center point of the 2D bounding box with corresponding depth. However, this method proved highly sensitive to sensor noise. Alternative approaches using color thresholding lacked robustness under varying lighting conditions and among different apple varieties. 
Using a segmentation model can robustly extract the apple pixels in a short time. 
In our segmentation model, with the structure shown in the right part of Fig.~\ref{fig:network_structure}, a high-resolution encoder extracts image features, which are processed by a dual-branch decoder: one branch produces a six-class semantic map (apple, canopy, branch, sky, ground, and other), and the other generates instance masks guided by prompts derived from detection boxes. 
The model was trained on our proprietary dataset, collected in different orchards in California and Michigan, such that the robot can adapt well to the field condition in the orchard we test.

\subsubsection{3D Localization With Point Cloud Clustering}\label{sec:localization}
To enhance robustness, we introduced a point cloud refinement step based on DBSCAN (Density-Based Spatial Clustering of Applications with Noise), an unsupervised clustering algorithm widely used for noise-resilient spatial clustering.~\citep{ester1996density} DBSCAN groups data points based on density, identifying clusters as areas of high point density separated by regions of lower density. It does not require prior knowledge of the number of clusters and is particularly effective at handling outliers—making it well-suited for real-world orchard environments where sensor noise and occlusions are common.

\begin{figure}[htbp]
    \centering
    \begin{subfigure}[b]{0.2\textwidth}
        \centering
        \includegraphics[width=1\textwidth]{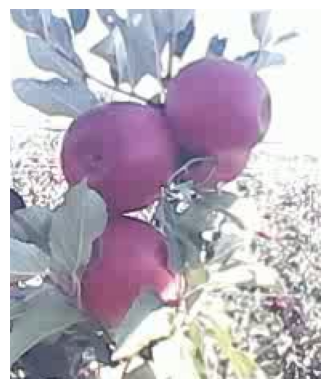}
        \caption{}
        \label{fig_perception_original}
    \end{subfigure}
    \hspace{5pt}
    \begin{subfigure}[b]{0.2\textwidth}
        \centering
        \includegraphics[width=1\textwidth]{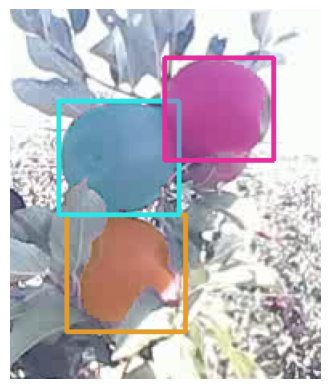}
        \caption{}
        \label{fig_perception_segmentation}
    \end{subfigure}
    \hspace{15pt}
    \begin{subfigure}[b]{0.45\textwidth}
        \centering
        \includegraphics[width=1\textwidth]{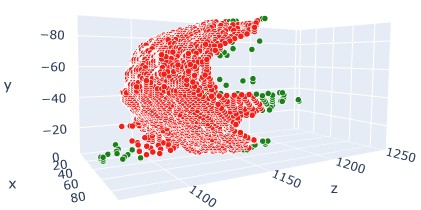}
        \caption{}
        \label{fig_perception_dbscan}
    \end{subfigure}
    \caption{The sample output of our perception algorithm. (a) The original image. (b) The detection and segmentation result. (c) The clustering-based depth estimation for one apple (the blue bounding-box in (b)). With a raw RGB-D image, the first thing is detect and segment the apples in the scene. Utilizing the pixel-wise segmentation, combined with the depth image, the point cloud for the apple can be retrieved. Using the DBSCAN, the noise points can be removed and the apple's position can then be estimated.}
    \label{fig_perception_sample}
\end{figure}

The sample output of our perception algorithm is shown in Fig.~\ref{fig_perception_sample}. 
Specifically, given aligned RGB and depth images from the ToF camera (Fig.~\ref{fig_perception_original}), the apples are firstly detected and segmented by our modules (Fig.~\ref{fig_perception_segmentation}), and then we can utilize the segmentation mask to extract the apple-related points from the point cloud. Using DBSCAN, we can cluster the extracted points and select the cluster that contains most points as the apple cluster, and then estimate the depth $z_d$ using the centroid of this cluster (Fig.~\ref{fig_perception_dbscan}). Finally, the 3D position of the apple is given by $\frac{z_d}{z_c}\cdot(x_c,y_c,z_c)$, where $(x_c,y_c,z_c)$ is the original 3D point corresponding to the center of the bounding-box after detection. This clustering-based method improves localization accuracy by leveraging the geometric structure of apple point clouds. Compared to the active laser-scanning approach used in our previous work, this solution significantly reduces processing time while maintaining comparable accuracy, thus improving overall harvesting efficiency.

\subsection{Arm Control}\label{sec:single_arm_control}
Once a target apple within the robot’s workspace is identified, the corresponding manipulator must move to the fruit’s location to execute the harvesting action. This subsection details the control framework for a single robotic arm, including the kinematic modeling, trajectory generation algorithm, and a robust trajectory tracking controller designed to ensure accurate motion execution under real-world disturbances.

\subsubsection{Kinematics Analysis}
\begin{figure}[htbp]
    \centering
    \includegraphics[width=0.72\linewidth]{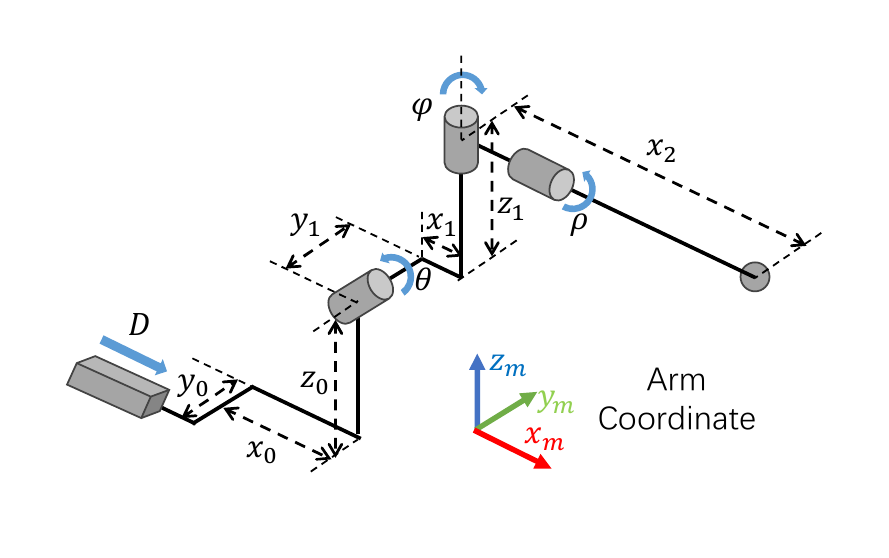}
    \caption{The kinematic model of one 4-Degree-of-Freedom robotic arm.}
    \label{fig_arm_kinematic_model}
\end{figure}

Consider an apple detected and localized by the perception algorithm described in Sec.~\ref{sec:localization}, with its position in the arm’s coordinate frame denoted as $p_a = [x_a, y_a, z_a]^T$. The first step is to determine whether the apple lies within the reachable workspace of the manipulator. This requires kinematic analysis.
The kinematic model of a single manipulator is illustrated in Fig.~\ref{fig_arm_kinematic_model}, and the corresponding parameters for both arms are provided in Table~\ref{tab:arm_parameters}. Let the end-effector position be represented as $p_e = [x_e, y_e, z_e]^T$. The forward kinematics of the arm can then be derived as follows:

\begin{equation}
    p_e=\left[
    \begin{array}{c}
         x_e \\
         y_e \\
         z_e
    \end{array}
    \right]
    =\left[
    \begin{array}{c}
        x_0+x_1\cos{\theta}-z_1\sin{\theta}+x_2\cos{\theta}\cos{\varphi}+D \\
        y_0+y_1-x_2\sin{\varphi} \\
        z_0+x_1\sin{\theta}+z_1\cos{\theta}+x_2\cos{\varphi}\sin{\theta}
    \end{array}
    \right]
\end{equation}

The inverse kinematics can then be derived accordingly. Given the target apple position, the corresponding joint values are computed as $q_a = [D_a, \theta_a, \varphi_a]^T$. These values are then checked against the joint limits specified in Table~\ref{tab:arm_parameters} to determine whether the apple is within the arm’s reachable workspace. Apples located outside of the feasible workspace are filtered out in the subsequent planning procedure.

\begin{table}[htbp]
    \centering
    \begin{tabular}{|c|c|c|}
        \hline
        Parameter & Arm 1 & Arm 2 \\
        \hline
        $x_0$ & 0.147 & 0.180 \\
        \hline
        $y_0$ & 0.017 & -0.023 \\
        \hline
        $z_0$ & 0.083 & 0.083 \\
        \hline
        $x_1$ & 0 & 0 \\
        \hline
        $y_1$ & 0.093 & -0.088 \\
        \hline
        $z_1$ & 0.138 & 0.140 \\
        \hline
        $x_2$ & 0.90 & 0.89 \\
        \hline
        $[D_{min},D_{max}]$ & [-0.02m,0.6m] & [-0.02m,0.6m] \\
        \hline
        $[\theta_{min},\theta_{max}]$ & [-17$^\circ$,30$^\circ$] & [-15$^\circ$,30$^\circ$] \\
        \hline
        $[\varphi_{min},\varphi_{max}]$ & [-19$^\circ$,19$^\circ$] & [-17$^\circ$,17$^\circ$] \\
        \hline
    \end{tabular}
    \caption{The kinematic parameters of the two arms}
    \label{tab:arm_parameters}
\end{table}

\subsubsection{Trajectory Generation}
Given the target apple position $p_a$, the next step is to generate a time-continuous reference trajectory that moves the manipulator from its current position to the target. Let the current end-effector position in Cartesian space be denoted as $p_0 = [x_0, y_0, z_0]^T$, which can be obtained using forward kinematics from the current joint configuration $q_0 = [D_0, \varphi_0, \theta_0]^T$. The corresponding Cartesian velocity of the manipulator can be derived as:
\begin{equation}\label{eq_jacobian}
    \left\{
    \begin{array}{ccl}
       \dot{x} & = & -\dot{\theta}(x_1\sin{\theta}+z_1\cos{\theta+x_2\sin{\theta}\cos{\varphi})}-\dot{\varphi}x_2\cos{\theta}\sin{\varphi}+\dot{D} \\
       \dot{y} & = & -\dot{\varphi}x_2\cos{\varphi} \\
       \dot{z} & = & \dot{\theta}(x_1\cos{\theta}-z_1\sin{\theta}+x_2\cos{\varphi}\cos{\theta})-\dot{\varphi}x_2\sin{\varphi}\sin{\theta}
    \end{array}
    \right.
\end{equation}

We adopt a quintic (5th-order) spline interpolation algorithm to generate smooth reference trajectories. For brevity, we illustrate the trajectory generation process using the $x$-coordinate as an example; the same method is applied to the $y$ and $z$ coordinates.

Let ${t, p_{x,r}(t), v_{x,r}(t), a_{x,r}(t)}$ represent the interpolated trajectory along the $x$-axis, where $p_{x,r}(t)$ is the reference position at time $t$, $v_{x,r}(t)$ is the reference velocity, and $a_{x,r}(t)$ is the reference acceleration. The trajectory is formulated as:
\begin{equation}\label{eq_traj_gen}
    \left\{
    \begin{array}{rcl}
        p_{x,r}(t) & = & k_5t^5+k_4t^4+k_3t^3+k_2t^2+k_1t+k_0 \\ 
        v_{x,r}(t) & = & 5k_5t^4+4k_4t^3+3k_3t^2+2k_2t+k_1 \\
        a_{x,r}(t) & = & 20k_5t^3+12k_4t^2+6k_3t+2k_2
    \end{array}
    \right.
\end{equation}

Consider a ``raw'' reference trajectory ${t_i, p_x(t_i)}$ consisting of $n$ discrete position-time pairs, where $n$ is a small integer. The trajectory begins at $t_1 = 0$ with $p_x(t_1) = x_0$. Additionally, the initial velocity $\dot{x}_0$ can be computed from Eq.~\eqref{eq_jacobian}. Using this information, we can augment the raw trajectory to include velocity and acceleration, resulting in the extended form ${t_i, p_x(t_i), v_x(t_i), a_x(t_i)}$ by:
$$v_x(t_i)=
\left\{
\begin{array}{cl}
     \frac{p_x(t_{i+1})-p_x(t_{i-1})}{t_{i+1}-t_{i-1}} & ,i\neq1,n \\
     \dot{x}_0 & ,i=1 \\
     0 & ,i=n
\end{array}
\right.
\,\,,\,\,\,
a_x(t_i)=
\left\{
\begin{array}{cl}
     \frac{v_x(t_{i+1})-v_x(t_{i-1})}{t_{i+1}-t_{i-1}} & , i\neq1,n \\
     0 & , \text{otherwise}
\end{array}
\right.
$$

With the augmented raw trajectory, the spline parameters can be efficiently computed using linear algebra. The quintic spline interpolation ensures that the resulting reference trajectory is smooth and dynamically feasible for the robotic arm to execute. This method supports both simple point-to-point movements and more complex scenarios where a high-level trajectory is defined by a sparse set of waypoints and associated timestamps.
The computational efficiency of the algorithm enables real-time responsiveness. For example, generating a trajectory from a high-level plan with 15 waypoints takes less than 0.1 seconds. This allows the arm to adapt its goal mid-motion, making it feasible to track and harvest moving apples.

\subsubsection{Robust Trajectory Tracking Control}
With the reference trajectory generated, the final step is to design a controller that enables the manipulator to accurately follow it. We implement a robust control strategy that regulates the end-effector to track the reference trajectory ${t, p_r(t), v_r(t)}$, where $p_r(t) = [p_{x,r}(t), p_{y,r}(t), p_{z,r}(t)]$ is the reference position and $v_r(t) = [v_{x,r}(t), v_{y,r}(t), v_{z,r}(t)]$ is the reference velocity.
The reference acceleration $a_r(t)$ is omitted, as the manipulator's servo motors are controlled via first-order velocity inputs $\dot{q} = [\dot{D}, \dot{\theta}, \dot{\varphi}]$. At time $t$, the tracking error is defined as $\varepsilon = [\varepsilon_x, \varepsilon_y, \varepsilon_z]$, where:
\begin{equation}
    \left\{
    \begin{array}{ccc}
        \varepsilon_x & = & x(t)-x_r(t) \\
        \varepsilon_y & = & y(t)-y_r(t) \\
        \varepsilon_z & = & z(t)-z_r(t)
    \end{array}
    \right.
\end{equation}

The trajectory tracking controller is then designed as:
\begin{equation}
    \left\{
    \begin{array}{ccl}
        \dot{\varphi} & = & (k_1\varepsilon_y-v_{y,r}+\eta_y)/x_2\cos{\varphi} \\
        \dot{\theta} & = & (-k_2\varepsilon_z+x_2\dot{\varphi}\sin{\theta}\sin{\varphi}+v_{z,r}-\eta_z)/(x_2\cos{\theta}\cos{\varphi}+x_1\cos{\theta}-z_1\sin{\theta}) \\
        \dot{D} & = & -k_x\varepsilon_x+x_2\dot{\theta}(\sin{\theta}\cos{\varphi}+x_1\sin{\theta}+z_1\cos{\theta})+x_2\dot{\varphi}\cos{\theta}\sin{\varphi}+v_{x,r}-\eta_x
    \end{array}
    \right.
\end{equation}

where $k_x,k_y,k_z>0$ are constant gains, and $\eta_x,\eta_y,\eta_z$ are computed by:
\begin{equation}
    \left\{
    \begin{array}{ccc}
        \eta_x & = & t_x\int_0^t[\varepsilon_x(\tau)+\mathrm{sgn}(\varepsilon_x(\tau))]d\tau \\
        \eta_y & = & t_y\int_0^t[\varepsilon_y(\tau)+\mathrm{sgn}(\varepsilon_y(\tau))]d\tau \\
        \eta_z & = & t_z\int_0^t[\varepsilon_z(\tau)+\mathrm{sgn}(\varepsilon_z(\tau))]d\tau 
    \end{array}
    \right.
\end{equation}

The constants $t_x$, $t_y$, and $t_z$ are positive scalar gains, and the sign function is defined as $\text{sgn}(x) = 1$ if $x \geq 0$, and $\text{sgn}(x) = -1$ if $x < 0$. This controller is designed based on principles of Lyapunov stability and robust control. It enables faster and more accurate arm motion, and ensures that the tracking error remains bounded and converges to zero asymptotically. A rigorous proof of stability can be derived following the methodology outlined in~\cite{Xian2004TAC}.

\subsection{Failure-Aware Coordination for Dual-Arm}\label{sec:coordination}
The dual-arm configuration necessitates an effective coordination strategy to ensure efficient and reliable apple harvesting. Upon receiving apple picking positions from the localization algorithm discussed in Sec.~\ref{sec:localization}, the system must assign and sequence apples to each arm in a way that prevents mutual interference and avoids collisions with nearby apples, which could result in unnecessary fruit loss.
Using the trajectory control approach described in Sec.~\ref{sec:single_arm_control}, each arm executes a harvest cycle by approaching its assigned target. Given that both arms share a single vacuum source (as described in Sec.~\ref{sec:valve_system}), we propose a failure-aware coordination strategy that leverages real-time pressure sensor feedback. This strategy allows the system to dynamically react to harvest outcomes, enabling the arms to operate in a staggered but efficient manner without requiring separate vacuum systems.
The proposed coordination method is robust, energy-efficient, and significantly reduces idle time. By effectively responding to harvest failures and adjusting the vacuum system configuration accordingly, the strategy maximizes the benefits of the dual-arm setup while overcoming the constraints imposed by a shared vacuum system.

\begin{figure}[!htbp]
    \centering
    \includegraphics[width=0.5\linewidth]{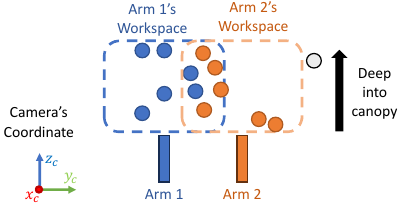}
    \caption{Illustration of the apple assignment algorithm. The top view of the joint workspace is shown in the figure. Apples located outside both workspaces are discarded initially. Apples within the overlapping region of the two workspaces are then assigned to either arm based on their $y$-coordinate. Subsequently, apples assigned to each arm are sorted in ascending order of their $z$-coordinate, which corresponds to their depth in the canopy. This ordering helps reduce the likelihood of collision between the arm and unharvested apples.}
    \label{fig:assignment}
\end{figure}

The apple assignment algorithm used in this work is a simplified version of the one presented in our previous study~\cite{lammers2024development}, and is outlined in Algorithm~\ref{alg_assignment}. In this context $dist_1(p_i)$ is the Euclidean distance from the end-effector to the apple positioned at $p_i$, thus $dist_1(p_i) = \infty$ indicates that apple $p_i$ lies outside the workspace of Arm 1, while $dist_1(p_i) < \infty$ signifies that the apple is within reach.
The algorithm begins by classifying apples into four categories based on their reachability: (1) reachable only by Arm 1, (2) reachable only by Arm 2, (3) reachable by both arms, and (4) unreachable by either arm (Lines 1–10).
Next, for apples that can be accessed by both arms, we sort them by their $y$-coordinates. Assuming there are $n$ apples that are reachable by either arm, the first $\frac{n}{2}$ (i.e., those closer to Arm 1) are assigned to Arm 1, and the remaining to Arm 2 (Lines 11–17).
Finally, the assigned apples for each arm are sorted in increasing order of their $z$-coordinates, which represent depth into the canopy (Line 18). This ensures that apples closer to the edge of the canopy are harvested first, reducing the risk of the arm colliding with unpicked apples in front of the target, and thereby minimizing unnecessary fruit loss.

\begin{algorithm}
    \renewcommand{\algorithmicrequire}{\textbf{Input:}}
	\renewcommand{\algorithmicensure}{\textbf{Output:}}
	\caption{Apple Assignment for Two Arms} 
	\label{alg_assignment} 
	\begin{algorithmic}[1]
		\REQUIRE Apple target positions $P=\{p_i\}$, where $p_i=\{x_i,y_i,z_i\}$ 
		\ENSURE Assigned targets $P_1=\{p_{i_{1}}\},\, P_2=\{p_{i_{2}}\}$
            \STATE Let $P_1=\{\},\, P_2=\{\},\, P_3=\{\}$
            \FORALL{$p_i\in P$}
                \IF{$dist_1(p_i)<\infty$ and $dist_2(p_i)=\infty$}
                    \STATE $P_1=P_1\bigcup\{p_i\}$
                \ELSIF{$dist_1(p_i)=\infty$ and $dist_2(p_i)<\infty$}
                    \STATE $P_2=P_2\bigcup\{p_i\}$
                \ELSE
                    \STATE $P_3=P_3\bigcup\{p_i\}$
                \ENDIF
            \ENDFOR
            \STATE Sort $P_3$ s.t. $y_{i_3}\leq y_{j_3},\forall i_3<j_3$
            \STATE $i_3=1$
            \WHILE{$|P_1|\leq\frac{|P_1|+|P_2|+|P_3|}{2}$}
                \STATE $P_1=P_1\bigcup\{p_{i_3}\}$, $P_3=P_3\backslash\{p_{i_3}\}$
                \STATE $i_3\leftarrow i_3+1$
            \ENDWHILE
            \STATE $P_2=P_2\bigcup P_3$
            \STATE Sort $P_1$ and $P_2$ s.t. $z_{i_1}\leq z_{j_1},\forall i_1<j_1$ and $z_{i_2}\leq z_{j_2},\forall i_2<j_2$
            \RETURN $P_1,\, P_2$
	\end{algorithmic} 
\end{algorithm}

Compared to the algorithm used in our previous work~\cite{lammers2024development}, the proposed simplified version introduces no major changes in logic but offers improved efficiency and reliability. Notably, it effectively avoids inter-arm collisions during the assignment stage, a common concern in multi-arm harvesting systems.
In addition to assignment-based coordination, the robot also incorporates a runtime safety check before each arm initiates movement toward a target apple. Specifically, the system verifies the current status of the other arm to ensure that executing the motion will not result in a collision. This real-time check, combined with the simplified assignment strategy, enhances the overall safety and robustness of the dual-arm system, allowing both arms to operate concurrently without risk of interference.

The next step is to drive the dual-arm system to harvest apples sequentially and efficiently. As introduced in our previous work~\citep{lammers2024development}, we employed a coordination strategy based on Temporal Logic to manage dual-arm operations. In this paper, with the addition of pressure sensors and the use of a single Time-of-Flight (ToF) camera as the primary sensing component, we synthesize an improved policy that enhances the system's overall efficiency.
To begin with, we partition each picking cycle into four discrete phases: \texttt{Approach}, \texttt{Detach}, \texttt{Retract}, and \texttt{Release}. Unlike in our previous implementation, the \texttt{Scanning} phase is omitted, as the new perception algorithm significantly reduces processing time while maintaining comparable localization accuracy, and imposes no constraints on the coordination logic. This finer partitioning enables more sophisticated scheduling strategies within each arm's operation.

Based on the physical configuration of the robot, two guiding principles are established:
(1) The two arms must not perform the \texttt{Detach} operation simultaneously to prevent a drop in vacuum pressure, which may occur when both arms are exposed to the atmosphere.
(2) If an apple detaches prematurely from an end-effector for any reason, the corresponding inlet valve should be closed immediately to minimize the impact on the other arm. This strategy leverages the fact that a properly attached apple creates a seal at the end-effector, thereby preserving the vacuum pressure for the second arm.

Considering the partitioned picking cycle and the two coordination principles previously discussed, we applied temporal logic to synthesize a policy that enables the dual-arm system to operate continuously—whenever apples are available—while strictly adhering to the constraints. Temporal Logic~\citep{baier2008principles} is a formal language that allows the specification of system goals and constraints over time using logical and temporal operators. The core operators include $\neg$ (Not), $\wedge$ (And), $\vee$ (Or), $\Rightarrow$ (Implies), $\Leftrightarrow$ (Equivalent), $U$ (Until), $\lozenge$ (Eventually), $\bigcirc$ (Next), and $G$ (Always). For a more in-depth treatment and practical examples, readers are referred to~\cite{kress2009temporal,zhao2023explore}.

In this work, we define two sets of atomic propositions:

\textbf{1. Environmental events}, denoted by $\varphi_e = \{Apple\_Detected_i, Apple\_Attached_i\}$, model the external state of the system:
\begin{itemize}
    \item $Apple\_Detected_i = \text{True}$ indicates that an apple has been detected and assigned to Arm $i$.
    \item $Apple\_Attached_i = \text{True}$ indicates that an apple is currently attached to the end-effector of Arm $i$, as inferred from pressure sensor feedback.
\end{itemize}

\textbf{2. Robot actions}, denoted by $\varphi_r = \{Approach_i, Retract_i, Open\_Valve_i\}$, model the robot's operational behavior:
\begin{itemize}
    \item $Approach_i = \text{True}$ indicates that Arm $i$ has completed its approach to the assigned apple.
    \item $Retract_i = \text{True}$ indicates that Arm $i$ has returned to its drop-off position.
    \item $Open\_Valve_i = \text{True}$ means the vacuum valve for Arm $i$ is open to the vacuum source, enabling attachment.
    \item Conversely, $\neg Open\_Valve_i = \text{True}$ implies that the valve is closed to the vacuum source and open to the atmosphere, allowing fruit release and preventing negative impact on the other arm's suction performance.
\end{itemize}

This formal representation enables us to encode both high-level coordination objectives and low-level safety constraints within a unified temporal logic framework, ensuring that the system can make autonomous decisions that are both efficient and reliable.

    

\begin{figure}[htbp]
	\centering
	\begin{subfigure}[b]{1.02\textwidth}
		\centering
		\includegraphics[width=1\textwidth]{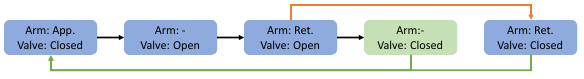}
		\caption{}
		\label{fig:single_task}
	\end{subfigure}
    
	\begin{subfigure}[b]{0.9\textwidth}
            \hspace{-45pt}
		\includegraphics[width=1.2\textwidth]{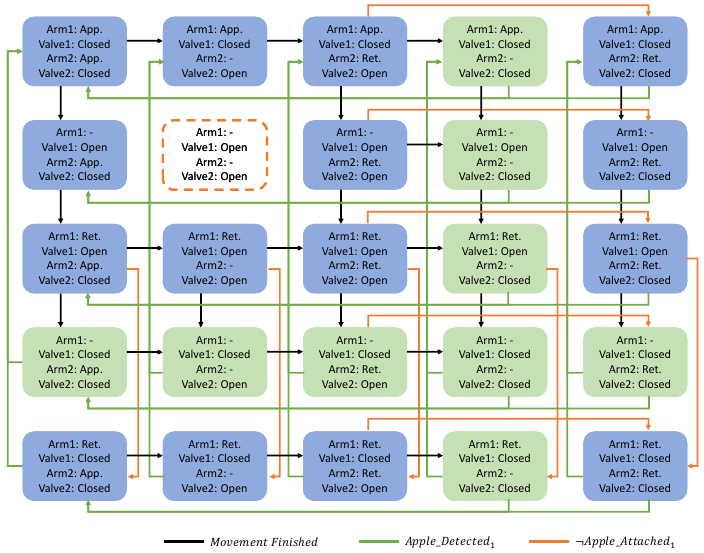}
		\caption{}
		\label{fig:dual_task}
	\end{subfigure}
    \caption{Illustration of (a) the workflow of a single arm and (b) the coordinated flow of the dual-arm system. 
    At the beginning of each cycle, the arm approaches (\textit{App.}) the apple. Once in position, the valve is opened to attempt attachment. Regardless of success or failure, the arm then retracts (\textit{Ret.}) to the drop-off position while maintaining suction. If the apple is successfully attached, the valve is closed at the drop-off position to release the apple. In the case of a failed attachment (orange line), the valve is closed immediately, skipping the release phase to minimize interference with the other arm. When a new apple is detected (green lines), a new harvesting cycle begins. 
    Green-highlighted states correspond to satisfaction of the goal condition $\varphi_{\text{goal}}$ (Eq.~\eqref{eq:goal}). 
    The red dashed state in (b) represents an invalid condition where both arms attempt attachment simultaneously, which is disallowed under the coordination constraint.}
    \label{fig:task}
\end{figure}

The primary objective of the system is to ensure that both arms continue harvesting as long as apples are assigned to them. This requirement can be formally expressed using temporal logic as:
\begin{equation}\label{eq:goal}
    \varphi_{goal}=G\lozenge(Retract_1\wedge Apple\_Attached_1)\wedge G\lozenge(Retract_2\wedge Apple\_Attached_2)
\end{equation}

This formula specifies that it should always eventually be the case that each arm returns to its drop-off position (\texttt{Retract}) with an apple successfully attached, indicating continuous and successful harvesting behavior for both arms.

Secondly, the system follows a predefined workflow for each arm. When an apple is assigned to Arm $i$, the arm first approaches the fruit, then opens its corresponding valve to initiate attachment. Regardless of whether the attachment is successful, the arm retracts to the drop-off position, closes the valve, and checks for the next available apple. This cycle enables the system to harvest continuously while remaining robust to occasional failures.
The workflow of each arm can be formally specified as:

\begin{equation}\label{eq_workflow_i}
    \varphi_{workflow,i}=\left\{
    \begin{array}{lr}
        \bigcirc Approach_i\Rightarrow (Apple\_Detected_i\wedge Retract_i) & \wedge \\
        \bigcirc Open\_Valve\Rightarrow Approach_i & \wedge \\
        Open\_Valve\Rightarrow \bigcirc Retract_i & \wedge \\
        (Retract_i\wedge Apple\_Attached_i)\Rightarrow\bigcirc(\neg Open\_Valve) & 
    \end{array}
    \right.
\end{equation}

In Eq.~\eqref{eq_workflow_i}, the first condition $\bigcirc Approach_i\Rightarrow (Apple\_Detected_i\wedge Retract_i)$
states that the arm may initiate the \texttt{Approach} action in the next step only if it has detected an apple and has completed its prior \texttt{Retract} action. Importantly, since \texttt{Apple\_Detected} can become true at any time (e.g., while the arm is still completing another cycle), this necessary condition ensures that the current cycle is not disrupted.
The second condition $\bigcirc Open\_Valve\Rightarrow Approach_i$ ensures that the valve is opened only after the approach phase has been completed. A necessary condition is used here to prevent simultaneous attachment attempts by both arms, thereby avoiding vacuum interference.
The third condition
$Open\_Valve\Rightarrow \bigcirc Retract_i$
dictates that after the valve is opened (and after a short wait), the arm must proceed to the drop-off location regardless of whether the attachment was successful. This rule maintains cycle robustness and prevents interference between the two arms.
The final condition
$(Retract_i\wedge Apple\_Attached_i)\Rightarrow\bigcirc(\neg Open\_Valve)$
specifies that if the arm reaches the drop-off position with an apple attached, the valve must be closed in the next step to release the apple.

Since both arms follow the same workflow structure, we define the overall system workflow constraint as:
\begin{equation}
    \varphi_{workflow}=\varphi_{workflow,1}\wedge\varphi_{workflow,2}
\end{equation}
This formulation ensures consistent and coordinated operation between the two arms, while enabling continuous, failure-resilient harvesting behavior.

Thirdly, it is necessary to explicitly specify the physical constraint that an apple can only be attached if the corresponding valve is open. This condition can be expressed using a necessary implication:
\begin{equation}
    \varphi_{attachment}=(Apple\_Attached_1\Rightarrow Open\_Valve_1)\wedge(Apple\_Attached_2\Rightarrow Open\_Valve_2)
\end{equation}
This formula ensures that apple attachment events are only valid when suction is actively supplied to the corresponding end-effector, reflecting the hardware limitation of the vacuum-based harvesting system.

Lastly, we consider the failure-aware coordination policy designed for the dual-arm system. Specifically, the system must ensure that at most one arm attempts apple attachment at a time, unless the other arm has already completed a successful attachment.
For example, if Arm~1 is attempting to harvest an apple (i.e., its valve is open), and Arm~2 has just reached the harvest position, Arm~2 should not open its valve simultaneously. Concurrent suction requests would compromise the vacuum strength, reducing the chances of successful detachment. However, once Arm~1 successfully attaches an apple—confirmed via pressure sensor feedback—it begins the \texttt{Retract} phase. At this point, Arm~2 is allowed to open its valve. Although Arm~1's valve remains open, the attached apple forms an effective seal, allowing sufficient pressure for Arm~2 to perform apple detachment.

In the case of harvest failure or accidental detachment during retraction, the valve should be closed immediately to minimize vacuum loss and avoid impacting the other arm’s operation. The following temporal logic expression captures the coordination constraints:
\begin{equation}
    \varphi_{coordination}=\left\{
    \begin{array}{ll}
       \neg Apple\_Attached_1\Rightarrow\bigcirc\neg Open\_Valve_1  & \wedge \\
       \neg Apple\_Attached_2\Rightarrow\bigcirc\neg Open\_Valve_2  & \wedge \\
       (Open\_Valve_1\wedge\neg Apple\_Attached_1)\Rightarrow\neg(\bigcirc Open\_Valve_2)  & \wedge \\
       (Open\_Valve_2\wedge\neg Apple\_Attached_2)\Rightarrow\neg(\bigcirc Open\_Valve_1)  &  
    \end{array}
    \right.
\end{equation}
This formulation ensures that the arms operate safely and efficiently under shared vacuum constraints, enabling robust coordination in both normal and failure conditions.

The overall objective and coordination strategy for the dual-arm harvesting system can be formally expressed by combining the previously defined constraints:
\begin{equation}
    \varphi=\varphi_{goal}\wedge\varphi_{workflow}\wedge\varphi_{attachment}\wedge\varphi_{coordination}
\end{equation}

Having specified the system objectives in temporal logic, we then convert the formula $\varphi$ into a directed graph representation, as illustrated in Fig.~\ref{fig:dual_task}. This graph provides an intuitive and structured depiction of the logical flow, enabling us to track how the system evolves over time in response to internal state changes and environmental events.
Starting from the initial node, the system can determine the next appropriate action based on the current status of both arms and sensor feedback. This approach allows the robot to execute behaviors that respect the specified constraints while reacting appropriately to failures and asynchronous events. As a result, efficient and reliable coordination of the dual-arm system is achieved through a formally grounded and interpretable control policy.

\begin{figure}[htbp]
    \centering
    \begin{subfigure}[b]{0.624\textwidth}
        \centering
        \includegraphics[width=1\textwidth]{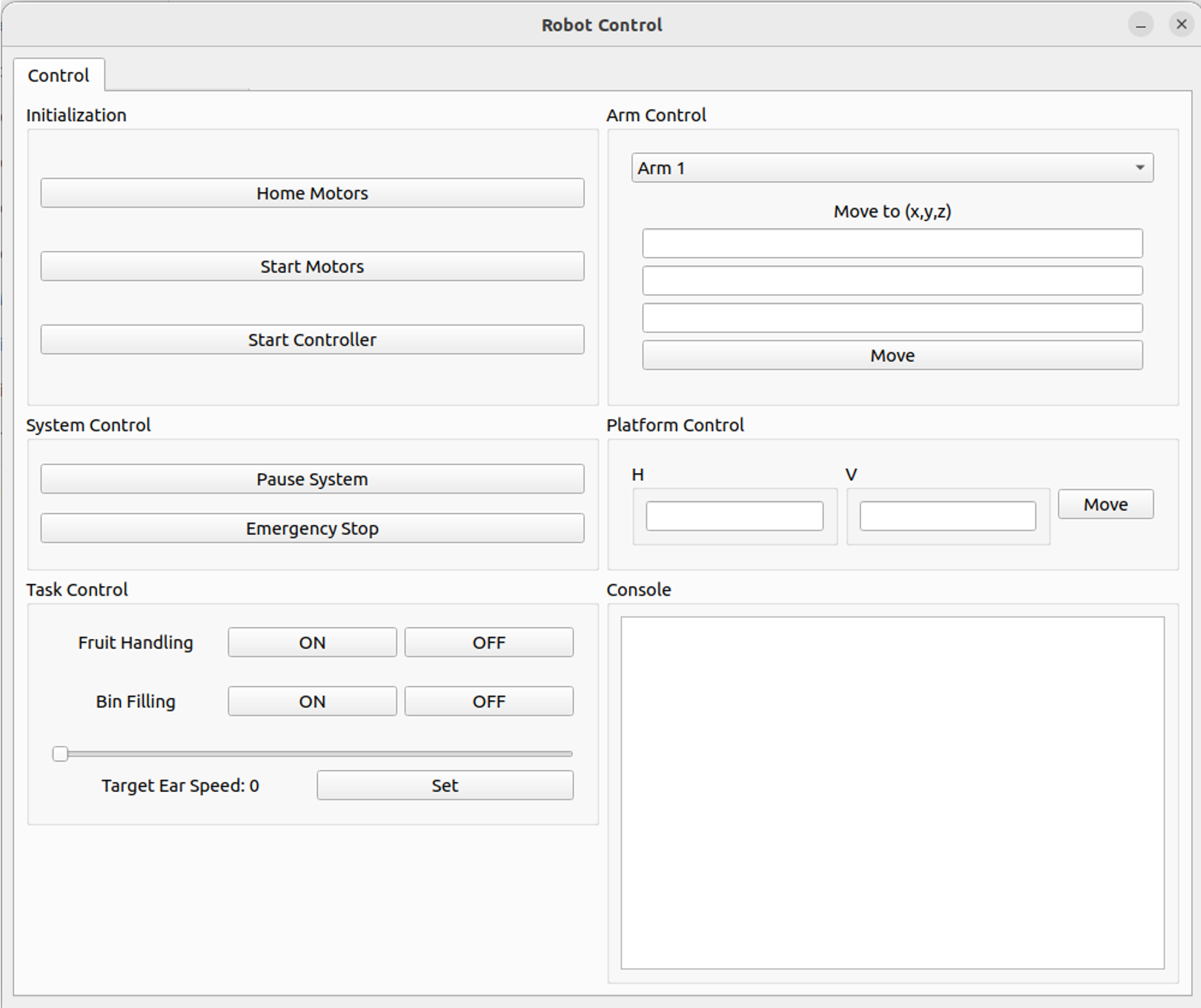}
        \caption{Control Panel}
        \label{fig_GUI_control}
    \end{subfigure}
    \begin{subfigure}[b]{0.32\textwidth}
        \centering
        \includegraphics[width=1\textwidth]{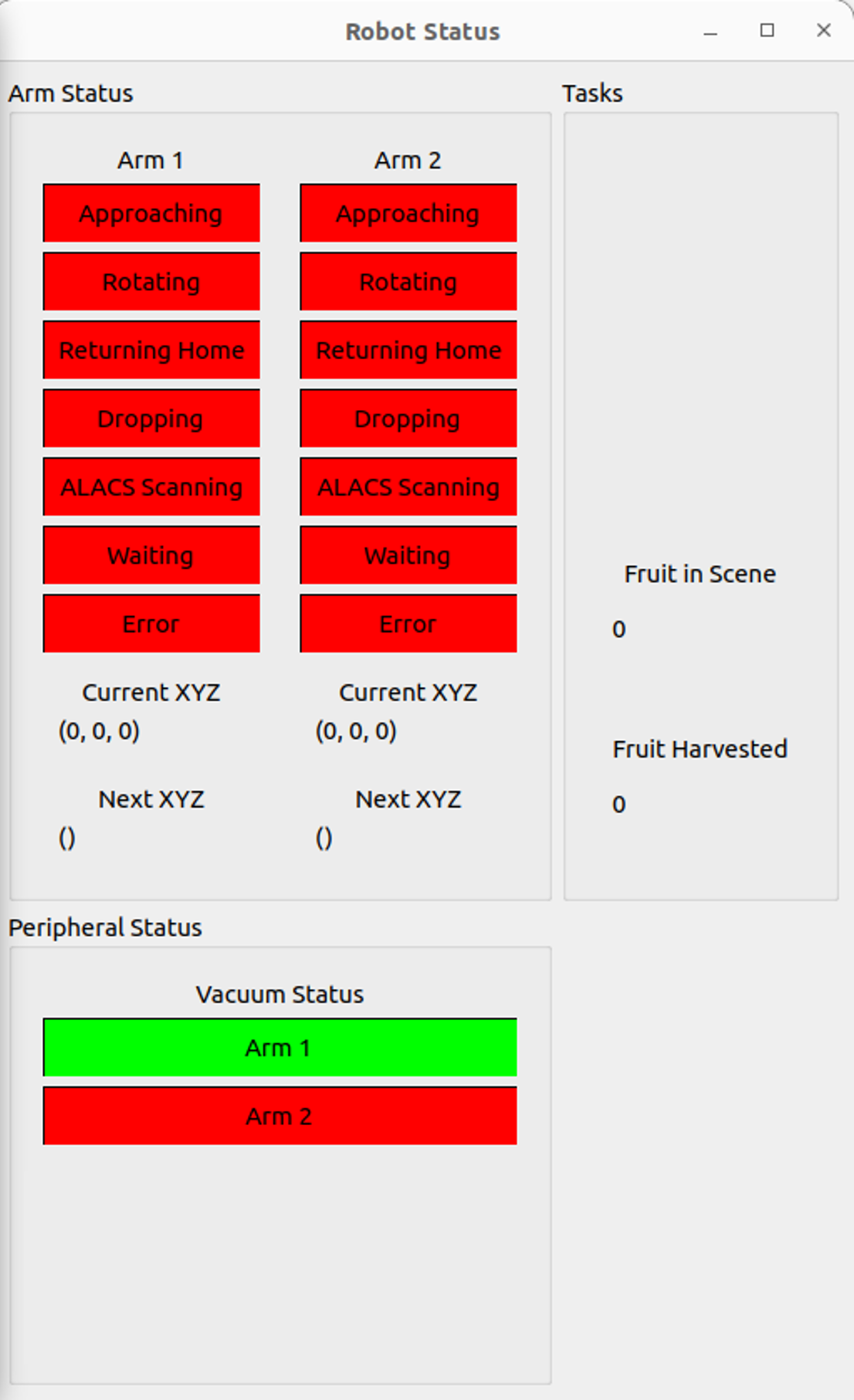}
        \caption{Status Panel}
        \label{fig_GUI_status}
    \end{subfigure}
    \caption{Graphical user interface (GUI) for the harvesting robot. (a) The control panel allows for system initialization, motion commands, control of the fruit handling system, and emergency stop functions. (b) The status panel displays real-time information such as the current task assigned to each arm, fruit statistics, and vacuum system status. The GUI assists operators in monitoring the harvesting process and enables timely manual intervention in case of system faults.}
    \label{fig_GUI}
\end{figure}

\subsection{Graphic User Interface}\label{sec:gui}
To facilitate field operation, a graphical user interface (GUI) was developed. As shown in Fig.~\ref{fig_GUI}, the interface consists of two main components. The first provides control functionalities, including system initialization, emergency stop, and manual movement commands. These features enhance operational convenience and allow for rapid manual intervention when necessary.
The second component displays real-time system information for monitoring during the harvesting process. The GUI not only streamlines testing procedures but also serves as a valuable tool for demonstrations and system validation. A more advanced version is currently under development, which will include expanded diagnostic capabilities such as point cloud visualization, localization outputs, and apple selection overlays on the camera feed. This enhanced interface will provide deeper insights into the harvesting process and support more efficient data collection for future analysis.

\section{Experiments}\label{sec:experiment}

\subsection{Field Evaluation}\label{sec:field_eval}
\begin{figure}[htbp]
	\centering
	\begin{subfigure}[b]{0.255\textwidth}
		\centering
		\includegraphics[width=1\textwidth]{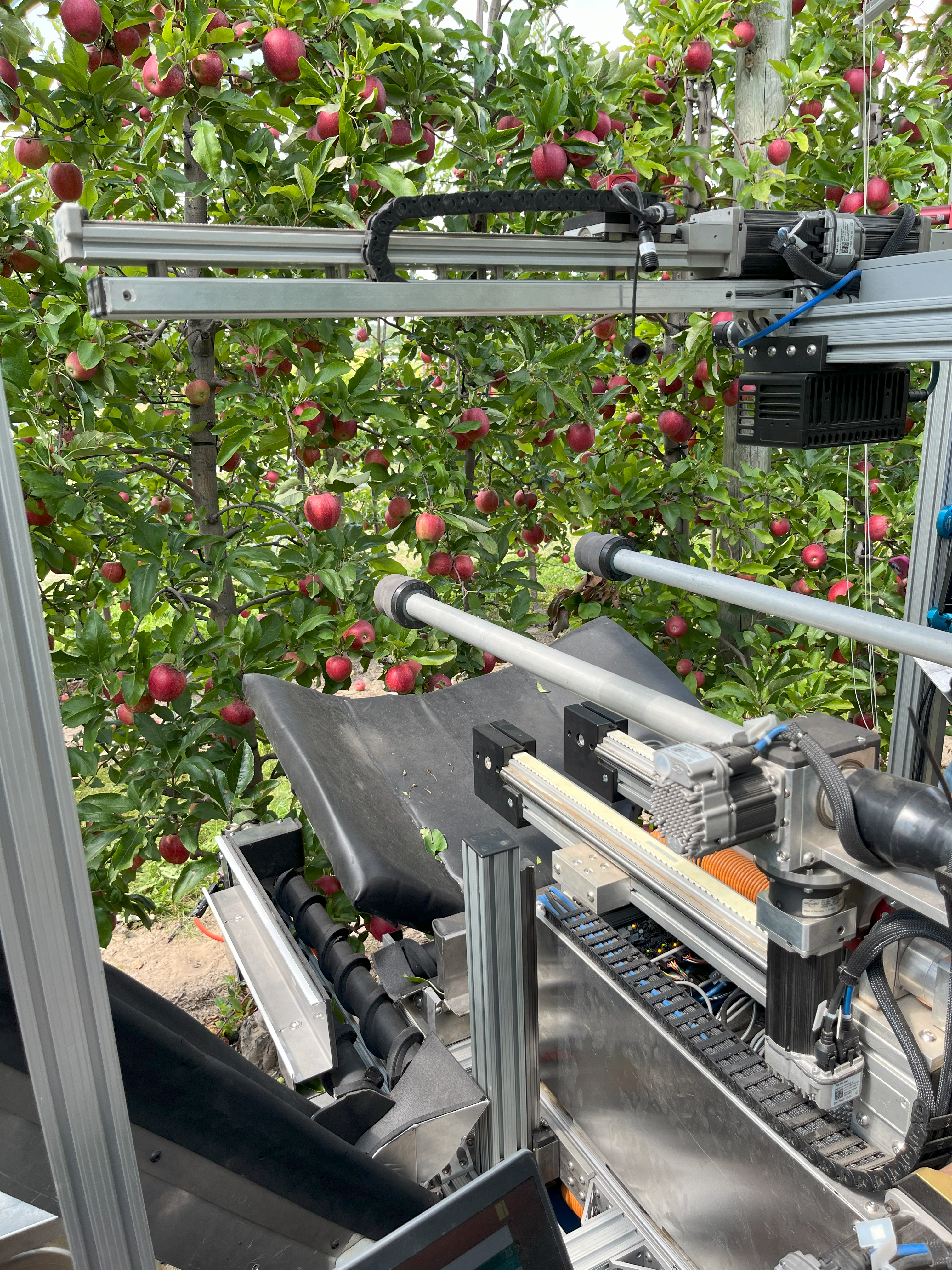}
		\caption{}
		\label{fig:orchard1}
	\end{subfigure}
    \hspace{10pt}
	\begin{subfigure}[b]{0.45\textwidth}
		\centering
		\includegraphics[width=1\textwidth]{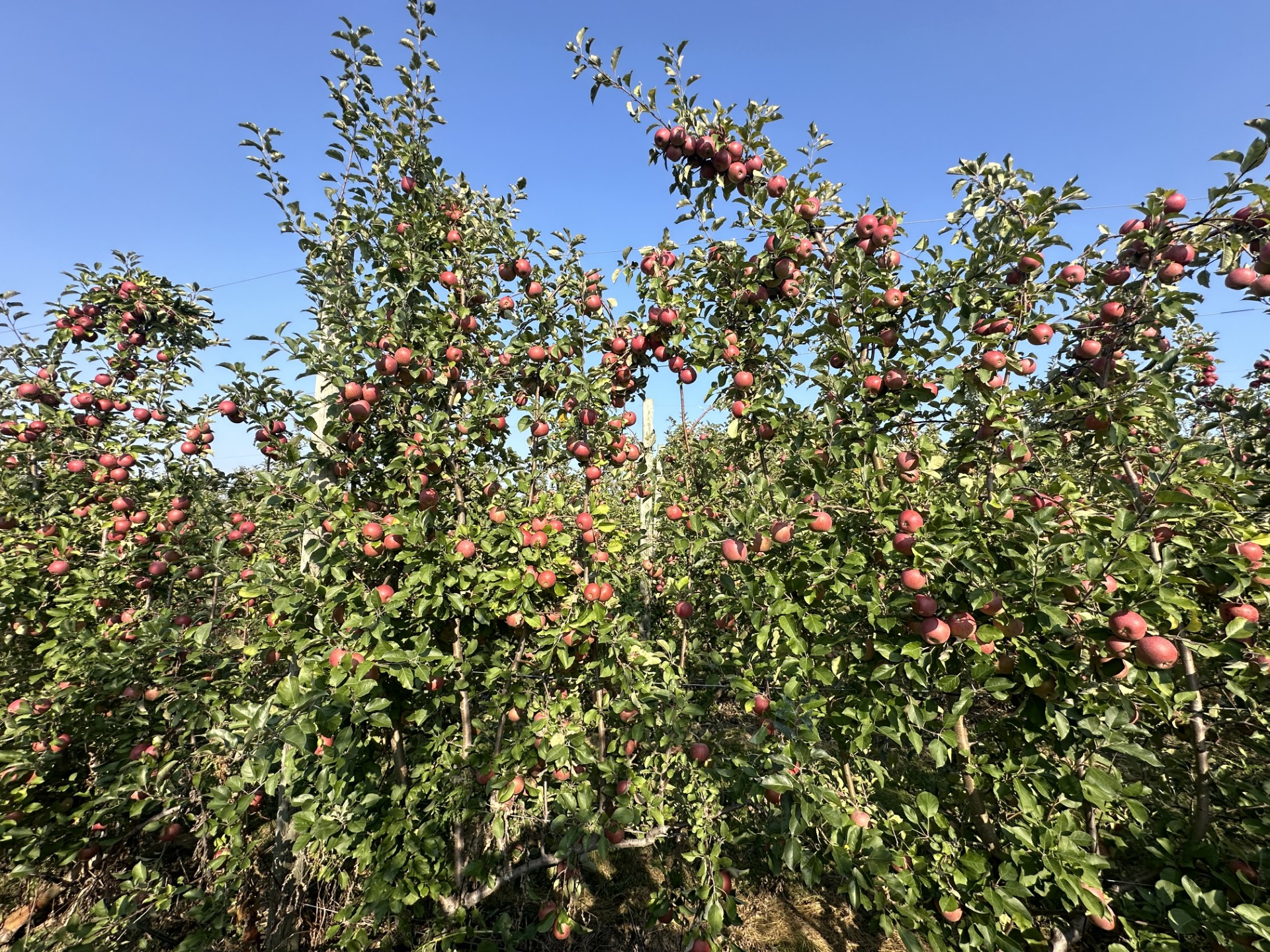}
		\caption{}
		\label{fig:orchard2}
	\end{subfigure}
\caption{Example images of apple trees from the two commercial orchards used for field evaluation.}
\label{fig:orchards}
\end{figure}

Field demonstrations of the dual-arm robotic harvesting system were conducted in two commercial orchards in Michigan, USA, during the 2024 harvest season. The orchard environments are shown in Fig.~\ref{fig:orchards}. The first demonstration took place on the morning of September 10, 2024, in a block of Gala apples, while the second was held on the afternoon of October 4, 2024, in a block of Fuji apples. Both varieties are bi-colored, with ripe apples exhibiting a predominantly red hue.
The first orchard employed a \textcolor{blue}{vertical fruiting-wall} structure, while the second used a UFO (Upright Fruiting Offshoots) structure. Both orchards featured average fruit density, with a mix of clustered and isolated apples, making them representative of typical commercial orchard conditions. As such, they provided suitable environments for validating the performance and robustness of the proposed robotic harvesting system.

The evaluation procedure was conducted as follows. The apple harvester was positioned facing the orchard canopy, and the tractor transported the robot along the orchard rows. The system currently operates in a stop-and-go mode: the tractor halts at each appropriate harvest location, after which the platform movement mechanism adjusts the robot’s position so that the dropping module makes contact with the canopy. This alignment maximizes the overlap between the robot’s workspace and the fruiting area.
Once the platform is in position, the autonomous harvesting procedure begins. After all accessible apples in the current position are either harvested or attempted, the operator terminates the harvesting cycle, and the tractor advances the robot to the next designated harvest location.

\subsection{Performance Analysis}\label{sec:performance_analysis}
The performance statistics from the 2024 field demonstrations are summarized in Table~\ref{tab:performance}. As shown, the success rates in the two orchards are 80.7\% and 79.7\%, respectively, demonstrating the effectiveness of our harvesting robot under different orchard structures.
Among the successfully harvested apples, 85.4\% and 88.6\% were picked on the first attempt in each orchard, highlighting the precision and robustness of the perception and control algorithms. It is important to note that some apples were heavily occluded by branches or foliage, rendering them effectively ``unharvestable''. In better-structured orchards with improved visibility and spacing, the robot is expected to achieve even higher performance.

\begin{table}[htbp]
    \centering
    \begin{tabular}{|c|c|c|c|c|}
        \hline
        Orchard\# & Attempted Apples & Success & 1st Attempt & $\geq$2 Attempts\\
        \hline
        1 & 322 & 260(80.7\%) & 222 & 38 \\
        \hline
        2 & 285 & 227(79.7\%) & 201 & 26 \\
        \hline
    \end{tabular}
    \caption{Summary of harvesting performance in two orchards, as measured by the number of attempted apples, successful harvests, and a breakdown of successes achieved on the first attempt versus those requiring two or more attempts.}
    \label{tab:performance}
\end{table}

\begin{figure}[htbp]
    \centering
    \includegraphics[width=0.7\linewidth]{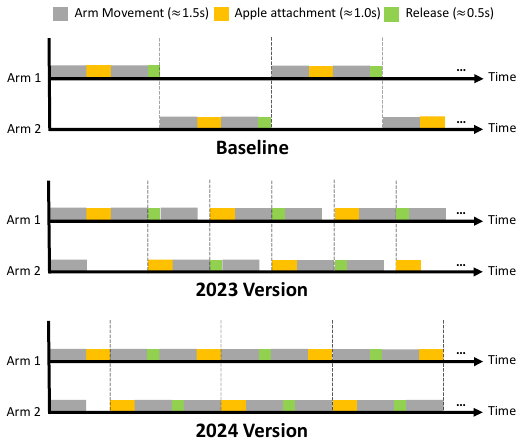}
    \caption{Cycle time analysis comparing the baseline (the dual-arm version without coordination), the 2023 dual-arm version, and the proposed 2024 version. Arm movement time is assumed to be identical across all cases, and timing values are based on averages from in-field operation. For the 2023 version, the average cycle time per apple corresponds to the interval between two dashed lines. In the 2024 version, both arms complete a cycle within that same interval, resulting in an effective per-apple cycle time equal to half the dashed-line spacing.}
    \label{fig:time_analysis}
\end{figure}

As the picking cycle time in real-world scenarios is highly influenced by the spatial distribution of apples, we first performed a theoretical analysis to estimate cycle duration under idealized conditions. The time breakdown is illustrated in Fig.~\ref{fig:time_analysis}. In each harvest cycle, the arm performs 4 sequential actions: approaching the apple, attaching it via vacuum suction and arm rotation, retracting, and releasing the fruit.

For simplicity, we assume that the time required for the approach phase is consistent across targets. Each of the four steps is assigned a time estimate based on the average values observed during field operation. Since the arm travels the same distance when approaching and retracting, these two phases are grouped under the label ``Arm Movement'' in the analysis, with identical time allocations.
In this analysis, we compare three coordination strategies: a baseline version, the 2023 version from our previous work~\citep{lammers2024development}, and the proposed 2024 version. The comparison is illustrated in Fig.~\ref{fig:time_analysis}.
The Baseline version adopts a no-coordination strategy, where the two arms operate strictly sequentially—only one arm is active at a time. Each arm executes the four steps (approach, attach, retract, release) in sequence, and the second arm begins only after the first has completed its full cycle. The average cycle time is represented by the interval between two neighboring dashed lines, approximately 4.5 seconds per apple.
The 2023 Version introduces a coordination strategy that allows parallel movement without vacuum interference. Specifically, both arms can perform the approach phase simultaneously. However, attachment and retraction phases are not allowed to run in parallel, as simultaneous execution would reduce the vacuum force available to each arm, thereby compromising picking reliability. Once one arm completes its release phase, the other arm is permitted to proceed with attachment. As shown in Fig.~\ref{fig:time_analysis}, this results in overlapping but partially staggered execution. The average cycle time per apple in this version is reduced to approximately 2.5 seconds—the interval between successive releases.
The 2024 Version, described in Sec.\ref{sec:coordination}, incorporates pressure sensor feedback and a new coordination algorithm to enable more parallel behavior. In this version, only the simultaneous execution of attachment is prohibited to ensure sufficient vacuum force. For example, in the first dashed-line interval in the 2024 timeline of Fig.\ref{fig:time_analysis}, the arms stagger attachment phases, but operate independently for all other steps. This coordination ensures that both arms can complete their cycles without mutual blocking. As a result, two releases occur within one cycle period, effectively halving the per-apple cycle time to approximately 2.25 seconds.

In the field demonstrations, the average cycle time per attempt for the dual-arm system was 5.97 seconds, while the average for a single arm was 8.29 seconds. During testing, the arm movement speed was tuned to 60\% of its maximum capacity to stay conservative in order not to damage the orchard canopy.
It is important to note that the theoretical cycle time analysis in Fig.~\ref{fig:time_analysis} assumes ideal conditions: all apples are positioned at uniform distances from the arms, and every harvest attempt is successful. However, in real-world scenarios, some attempts result in harvest failure. In the 2023 Version, if one arm fails to attach its target apple, the other arm must still wait for the failed arm to complete its retraction, leading to avoidable delays.
By contrast, the 2024 Version—described in Sec.~\ref{sec:coordination}—leverages pressure sensor feedback and a reactive coordination strategy. When a harvest failure is detected, the system allows the other arm to begin attaching its assigned apple without delay. This adaptive mechanism significantly improves picking efficiency under practical field conditions.
Furthermore, the benefits of the improved coordination strategy are proportional to the arm movement speed. If the system were improved for operating at higher speeds, significant improvement for harvesting efficiency would be expected. As shown in the theoretical analysis, the average cycle time under the 2024 Version is approximately 50\% of that for the baseline, demonstrating the scalability of the strategy.

\subsection{Failure Analysis}\label{sec:failure_cases}
Harvesting in outdoor orchard environments presents numerous challenges, where multiple factors can jointly contribute to unsuccessful attempts. To identify the root causes of harvest failures for future system improvements, we conducted a failure analysis by reviewing video recordings from the field demonstrations. In total, 130 harvesting videos were examined.
The primary causes of failure were found to originate from two main sources: the perception algorithm and the vacuum system.

\begin{table}[htbp]
    \centering
    \begin{tabular}{|c|c|c|c|c|c|c|}
       \hline
       Source & \multicolumn{3}{|c|}{Perception Algorithm} & \multicolumn{3}{|c|}{Vacuum System} \\
       \hline
       \hline
       Cause  & Exposure & Occlusion & Cluster & Sealing & Detachment & Interference \\
       \hline
       Count  & 101 & 207 & 159 & 44 & 22 & 14 \\
       \hline
    \end{tabular}
    \caption{Summary of failure cases observed during field evaluation. For the perception algorithm category, some failures may involve multiple overlapping causes.}
    \label{tab:failure analysis}
\end{table}

For the perception algorithm, the primary contributors to harvest failures are as follows:
\begin{itemize}
    \item \textbf{Camera exposure issues:} Intense sunlight either reflecting off the surface of apples or directly entering the camera lens can cause severe color distortion. This significantly degrades the performance of both the detection and segmentation algorithms, leading to missed or inaccurate apple localization.
    \item \textbf{Occlusion:} Foliage and branches frequently obscure parts of the apple, reducing the number of visible pixels. More critically, occlusion can fragment the point cloud of a single apple into multiple disconnected clusters. Since our clustering-based depth estimation algorithm selects the cluster with the most points to determine apple depth, such fragmentation can result in incomplete data and inaccurate depth estimation.
    \item \textbf{Apple clusters:} Closely packed apples can lead to over-detection, where multiple bounding boxes are generated for a single fruit (e.g., three bounding boxes for two apples). This can introduce "pseudo" apple positions in the subsequent localization process described in Sec.~\ref{sec:localization}.
\end{itemize}

These challenges—whether occurring individually or in combination—often lead to significant localization errors, especially in unstructured orchard environments. To address them, future work will focus on integrating more advanced detection and segmentation models, enriching the training dataset with diverse orchard conditions, and improving the robustness of the depth estimation algorithm. Additionally, as orchard structures become more uniform and well-maintained, the system's performance is expected to improve accordingly.

For the vacuum system, the main factors contributing to harvest failures are as follows:
\begin{itemize}
    \item \textbf{Sealing failure:} In some cases, although the apple is accurately localized and the arm approaches correctly, leaves or thin branches obstruct the end-effector, preventing it from forming a proper seal with the fruit. As a result, the suction mechanism fails to attach the apple.
    \item \textbf{Insufficient detachment force:} This occurs when the end-effector successfully attaches to the apple but fails to detach it from the tree. The most common cause is excessive stem strength, often due to the unripeness of the fruit, resulting in a bond too strong for the vacuum system to overcome.
    \item \textbf{Holding interference:} After an apple is detached and the arm begins to return to the dropping location, it may collide with surrounding branches. Such interference can dislodge the apple from the end-effector, leading to harvest failure.
\end{itemize}

To mitigate these issues, several improvements can be pursued. A selective harvesting strategy could be employed to target only ripe apples with lower detachment resistance. Enhancing the perception system to detect not only apples but also surrounding canopy elements (e.g., branches and leaves) would enable better environmental awareness. Additionally, integrating a more sophisticated motion planning algorithm would allow the arms to avoid branch collisions during both approach and retraction phases, thereby reducing the likelihood of sealing and holding failures while minimizing fruit damage.


\section{Discussions}\label{sec:discussion}
Although the field tests conducted in the autumn of 2024 demonstrated the effectiveness and progress of our dual-arm robotic harvesting system, several limitations were also observed. To move toward commercial deployment, further improvements in both hardware design and software algorithms are necessary. 

Perception plays a critical role in autonomous apple harvesting. While our algorithm achieves high success rates under minimal or partial occlusion, heavily occluded apples remain a significant challenge. In such cases, dense foliage and branches can hinder detection entirely, even when the apple lies within the robot's reachable workspace. Moreover, heavy occlusion reduces the number of visible apple pixels, degrading the performance of the clustering-based localization algorithm. This not only leads to inaccurate depth estimation but also complicates the selection of an appropriate picking point. In particular, when only part of the apple is visible, the identified picking point may deviate significantly from the fruit’s center, lowering the chances of successful detachment even if localization is otherwise accurate.
To address these challenges, a new perception algorithm is under development that incorporates improved detection capabilities under heavy occlusion and more accurate localization using only partial fruit visibility. In addition, the active laser-scanning module used in our previous work~\citep{lammers2024development} may be selectively reintroduced for extreme cases, while the standard Time-of-Flight (ToF) localization remains in use for typical conditions. This hybrid strategy offers a balance between localization accuracy and harvesting efficiency.
To further improve the selection of picking points, visual servoing techniques~\citep{ren2024mobile,parsa2024modular}—which employ a camera mounted on the end-effector—can provide dynamically updated views as the robot moves, resulting in more precise fruit localization. Another promising direction involves segmenting visible apple pixels and estimating the occluded portion of the fruit, as explored in recent studies~\citep{magistri2024improving,gene2023looking}. With this approach, a more complete understanding of the apple's geometry can be achieved, enabling better picking point selection and higher success rates in challenging scenarios.

During operation, physical interference between the robot and the canopy can disrupt the picking process and raise concerns about both machine safety and potential damage to the trees. These interferences primarily originate from two components: the dropping module and the robotic arms.
The dropping module plays a crucial role in protecting apples during release. Its broad, foam-padded ramp design provides tolerance for variations in arm positioning, which reduces the risk of bruising and enhances overall picking efficiency. However, during platform movement, the extended structure of the dropping module may collide with protruding branches. Such contact can destabilize the apple tree, potentially causing fruit drop or, in more severe cases, damaging branches. Similar concerns arise as the vehicle transports the robot through the orchard.
To address these issues, a more compact and mechanically robust version of the dropping module is currently under development. Additionally, updates to the arm design will aim to reduce its overall mass, allowing faster and safer platform movement while maintaining high picking efficiency.
Branch collisions involving the arms can also lead to unintended apple loss and tree damage. Due to the arm’s compact design, certain apple positions permit only a single kinematically valid configuration, which may increase the likelihood of contact with the canopy. However, the direction and manner in which the arm approaches an apple significantly affect the outcome of such interactions. Some approaches may only displace foliage slightly, while others may cause breakage.
To mitigate these risks, a more sophisticated trajectory generation strategy is being developed. This strategy will incorporate deformation-aware planning, taking into account the physical response of branches to contact. Such a capability is expected to reduce canopy interference and enhance both safety and harvesting success.

In the current version of the robot, movement through the orchard still requires manual operation. A driver must operate the tractor to transport the system, frequently starting and stopping the vehicle at each harvesting position. This mode of operation is not energy-efficient and poses challenges in consistently positioning the robot optimally for apple picking.
As demonstrated in recent work~\citep{miao2023efficient,xiong2020autonomous}, an electric autonomous vehicle can effectively address this limitation. Such a system, when integrated with the harvesting robot, would allow continuous navigation through the orchard, with vehicle speed controlled by the main computer. Ideally, the robot would traverse the orchard at a slow, adaptive pace—adjusting velocity based on apple density—while the platform movement module repositions the harvesting arms according to fruit distribution. With lightweight arms capable of high-speed operation, apples could be harvested continuously without requiring the vehicle to stop.
This fully integrated setup would represent a major step toward truly autonomous apple harvesting and significantly advance the system’s readiness for commercial deployment.

\section{Conclusion}\label{sec:conclusion}
In this paper, a new design of dual-arm apple harvesting robot is presented. The system integrates a Time-of-Flight (ToF) camera, a dual-arm robotic harvesting module, a centralized vacuum system, and a fruit gathering system. Compared to the previous version, a platform movement mechanism that enhances the flexibility and adaptability of the robot within orchard environments, is introduced.
a novel foundation-model-based apple localization algorithm that delivers fast and robust performance under challenging orchard conditions is developed. Additionally, a new dual-arm coordination strategy that leverages sensor feedback to improve picking efficiency and reduce idle time associated with sharing a centralized vacuum source, is proposed.
Field evaluations were conducted in two commercial orchards in Michigan, USA, achieving success rates of 80.7\% and 79.7\%, with an average cycle time of 5.97 seconds per apple. These results validate the effectiveness and potential of the system.
Future work will focus on advancing the robot’s autonomy by incorporating an autonomous electric vehicle and developing a platform movement algorithm capable of dynamically selecting optimal harvesting positions based on fruit distribution. Also, the perception performance under extreme conditions such as overexposure, backlighting, and heavy canopy occlusion, should be enhanced. Finally, a new motion planning algorithms that minimize interaction with the canopy is under development, thus reducing both fruit and tree damage and further improving system robustness.
    
\vspace*{\fill}
	
\subsubsection*{Acknowledgements}
This project was funded by the USDA-SCRI project ``AIMS for Apple Harvest and In-Field Sorting'' (Project No. 2023-51181-41244). The findings and conclusions in this paper are those of the authors and should not be construed to represent any official USDA or U.S. Government determination or policy. Mention of commercial products in the paper does not imply endorsement by USDA over those not mentioned.

\bibliographystyle{apacite}
\bibliography{reference}
	
\newpage
	
	
\newpage
	
	
\end{document}